%% file: main.tex
\documentclass[preprint,12pt]{elsarticle}

\usepackage{amssymb}
\usepackage{amsmath}
\usepackage{booktabs}
\usepackage{tikz}
\usepackage{float}
\usepackage{multirow}
\usepackage{color}
\usepackage{xcolor}
\usepackage{hyperref}
\usepackage{array}
\usepackage{rotating} 

\AtBeginEnvironment{tabular}{\scriptsize}
\AtBeginEnvironment{tabularx}{\scriptsize}
\AtBeginEnvironment{longtable}{\scriptsize}

\makeatletter
\def\ps@pprintTitle{%
  \let\@oddhead\@empty
  \let\@evenhead\@empty
  \def\@oddfoot{\footnotesize\itshape\hfill\@date}%
  \let\@evenfoot\@oddfoot
}
\makeatother

\begin{document}

\begin{frontmatter}



\title{MoCHA-former: Moiré-Conditioned Hybrid Adaptive Transformer for Video Demoiréing}

\author[inst1]{Jeahun Sung}            
\ead{jhseong@cau.ac.kr}
    
\author[inst1]{Changhyun Roh}          
\ead{changhyunroh@cau.ac.kr}

\author[inst2]{Chanho Eom}             
\ead{cheom@cau.ac.kr}

\author[inst1]{Jihyong Oh\corref{cor1}}
\ead{jihyongoh@cau.ac.kr}

\affiliation[inst1]{%
  organization={Department of Imaging Science, Graduate School of Advanced Imaging Science, Multimedia \& Film, Chung-Ang University},
  addressline={},
  city={Seoul},
  postcode={06974},
  state={},
  country={South Korea}
}

\affiliation[inst2]{%
  organization={Department of Metaverse Convergence, Graduate School of Advanced Imaging Science, Multimedia \& Film, Chung-Ang University},
  addressline={},
  city={Seoul},
  postcode={06974},
  state={},
  country={South Korea}
}

\cortext[cor1]{Corresponding author \\ 
Full postal address: Graduate School of Advanced Imaging Science, Multimedia \& Film, Chung-Ang University, Seoul, 06974, Korea.}


\makeatletter
\g@addto@macro\elsaddress{%
  \begin{center}\url{https://cmlab-korea.github.io/MoCHA-former/}\end{center}%
}
\makeatother


\begin{abstract}
Recent advances in portable imaging have made camera-based screen capture ubiquitous. Unfortunately, frequency aliasing between the camera’s color filter array (CFA) and the display’s sub-pixels induces moiré patterns that severely degrade captured photos and videos. Although various demoiréing models have been proposed to remove such moiré patterns, these approaches still suffer from several limitations: (i) spatially varying artifact strength within a frame, (ii) large-scale and globally spreading structures, (iii) channel-dependent statistics and (iv) rapid temporal fluctuations across frames. We address these issues with the Moiré Conditioned Hybrid Adaptive Transformer (MoCHA-former), which comprises two key components: Decoupled Moiré Adaptive Demoiréing (DMAD) and Spatio-Temporal Adaptive Demoiréing (STAD). DMAD separates moiré and content via a Moiré Decoupling Block (MDB) and a Detail Decoupling Block (DDB), then produces moiré-adaptive features using a Moiré Conditioning Block (MCB) for targeted restoration. STAD introduces a Spatial Fusion Block (SFB) with window attention to capture large-scale structures, and a Feature Channel Attention (FCA) to model channel dependence in RAW frames. To ensure temporal consistency, MoCHA-former performs implicit frame alignment without any explicit alignment module. We analyze moiré characteristics through qualitative and quantitative studies, and evaluate on two video datasets covering RAW and sRGB domains. MoCHA-former consistently surpasses prior methods across PSNR, SSIM, and LPIPS.
\end{abstract}

\begin{keyword}
Video demoiréing \sep
RAW domain \sep
ISP \sep
Vision Transformer
\end{keyword}

\end{frontmatter}



\input{sections/introduction}

\input{sections/RelatedWorks}

\input{sections/ProposedMethod}

\input{sections/Experiments}

\input{sections/Conclusion}

\input{sections/Acknowledgements}







\bibliographystyle{elsarticle-num} 
\bibliography{ref} 

\end{document}

%% file: sections/introduction.tex
\section{Introduction} \label{sec:intro}

In recent years, the rapid development of mobile digital cameras leads to the widespread adoption of screen capturing as a common content-sharing method. However, capturing screens with a camera often introduces severe moiré patterns, which significantly degrade visual quality. These moiré artifacts are mainly caused by frequency aliasing between the camera's color filter array (CFA) and the sub-pixel structure of the screen \cite{mopnet,rrid, esdnet, tpami2023, tmm22, awudn}. Such distortions hinder the visibility of the underlying content and can further alter the overall color distribution, leading to a substantial degradation in the quality of the original image or video.

Recently, several learning-based demoiréing models~\cite{3,mopnet,5,6,7,8,9,dmcnn,11,awudn,13,tmm22,rrid,esdnet,vdemoire,rawvdemoire,dtnet,stdnet} have been proposed to remove moiré patterns. 
MopNet~\cite{mopnet} performs image demoiréing using a multi-scale architecture, while FHDe2Net~\cite{fhde2net} addresses demoiréing for FHD images.
However, when these image demoiréing \cite{mopnet, fhde2net} models are applied to video demoiréing task, they often suffer from degraded temporal consistency. To overcome this limitation, recent works \cite{vdemoire,rawvdemoire,dtnet} have proposed video demoiréing models. VDemoiré~\cite{vdemoire}, RawVDemoiré~\cite{rawvdemoire} and DTNet \cite{dtnet} achieve temporal consistency by employing PCD alignment~\cite{edvr} together with a multi-scale region-level relation loss. Despite these advancements in video demoiréing methods, existing video demoiréing models still face several challenging issues.
 
First, moiré patterns exhibit varying intensities and structures even within a single frame. This phenomenon occurs due to factors such as the camera’s perspective distortion, lens aberrations, and nonlinear distortions introduced by the ISP pipeline during the screen-capturing process \cite{tip2018,mopnet,moirepose}. Nevertheless, most existing models \cite{mopnet,5,6,7,8,9,dmcnn,11,awudn,13,tmm22,rrid,esdnet} apply uniform processing across all pixels regardless of moiré strength. As a result, not only are moiré patterns removed, but also the underlying content is often over-smoothed, compromising image fidelity \cite{dmcnn, mopnet}.

\begin{figure}[t]
    \centerline{\includegraphics[width=\columnwidth]{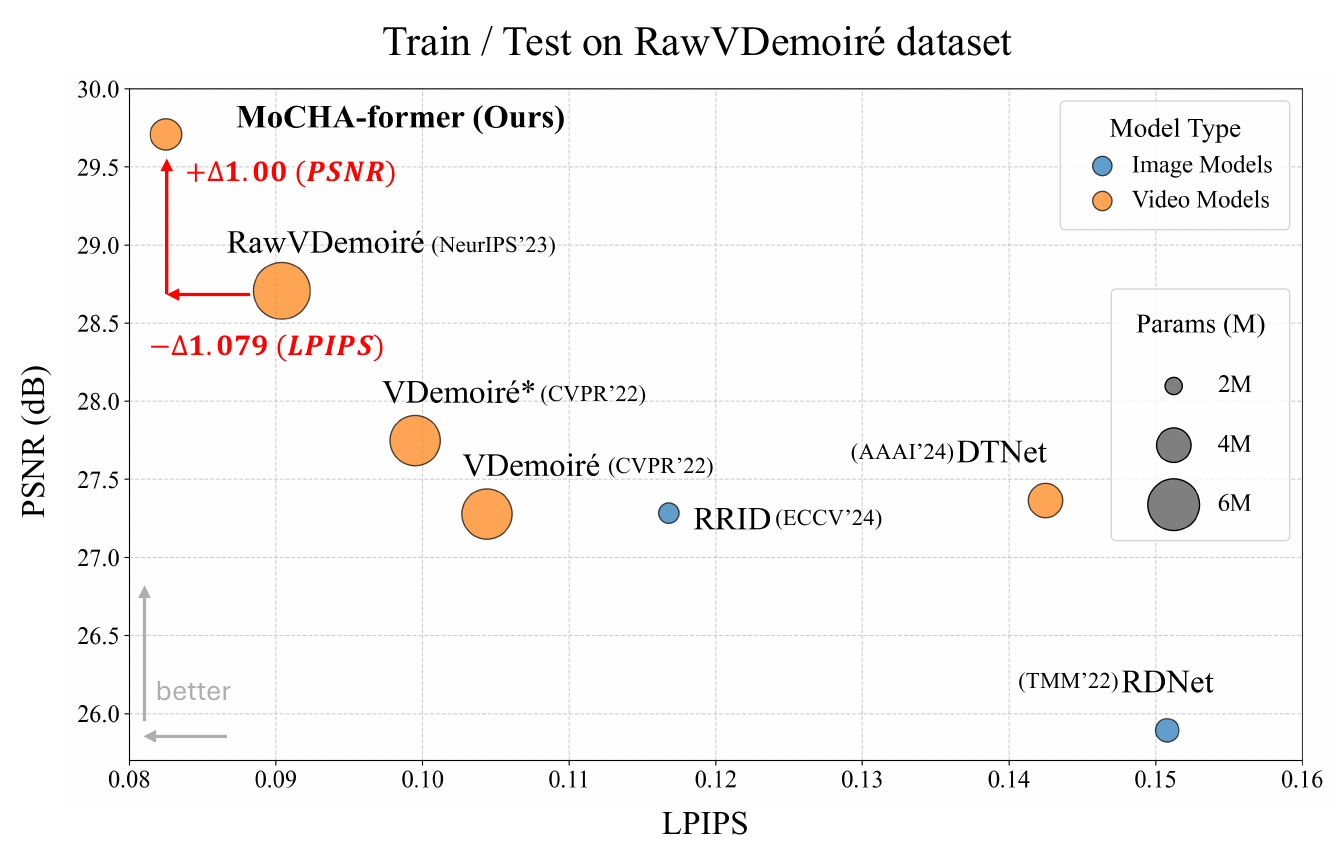}}
    \caption{Our MoCHA-former outperforms previous methods~\cite{rawvdemoire,vdemoire,rrid,dtnet,tmm22} in both PSNR$\uparrow$ and LPIPS$\downarrow$ while using fewer parameters.}
    \label{fig:psnr_lpips_param} 
\end{figure}

Second, most existing demoiréing methods \cite{3,mopnet,dmcnn,tmm22,vdemoire,rawvdemoire,dtnet} struggle to remove large-scale, globally distributed moiré patterns in both images and videos. While recent methods adopt multi-scale architectures \cite{3,mopnet,5,11} and symmetric encoder-decoder networks such as U-Net \cite{unet, awudn, rrid}, they still fail to fully capture the global characteristics of moiré artifacts. Although a transformer-based approach \cite{transformer} has been explored in the video demoiréing domain to exploit global information \cite{stdnet}, its sparse attention mechanism still results in heavier computational demand compared to CNN-based models \cite{vdemoire, rawvdemoire, dtnet}.

Third, existing RAW demoiréing models~\cite{tmm22,rawvdemoire} fail to effectively address the characteristics of moiré patterns in the RAW domain. In the RAW domain, moiré patterns exhibit distinct appearances across different channels. To exploit this property, RDNet~\cite{tmm22} and RawVDemoiré~\cite{rawvdemoire} attempt to model the channel-dependent nature of moiré patterns; however, their approaches are limited to considering these characteristics only in the spatial domain. Since moiré patterns can be more distinctly separated in the frequency domain (particularly in the Fourier domain) such properties are not fully leveraged by previous methods~\cite{tmm22,rawvdemoire}.

Fourth, image demoiréing models \cite{3,mopnet,dmcnn,tmm22} generally lack temporal modeling capabilities. When applied directly to video sequences, they often introduce flickering artifacts due to the absence of temporal consistency mechanisms \cite{vdemoire, rawvdemoire, dtnet, stdnet, fpanet}. Several recent methods incorporate temporal-aware modules to address this issue. For example, \cite{vdemoire, rawvdemoire, dtnet, fpanet} utilize the pyramid, cascading, and deformable convolution framework of EDVR \cite{edvr} for frame alignment. In addition, \cite{vdemoire} proposes a temporal consistency loss to encourage frame-to-frame coherence. However, these approaches typically involve significant parameter overhead and are prone to training instability \cite{basicvsr++,chan2021understanding}.

To analyze the aforementioned limitations, we provide both qualitative and quantitative evaluations for the aforementioned case. Specifically, our analysis considers the color distortion caused by moiré patterns, their spatially varying characteristics within a single image, their abrupt temporal variations, their channel-dependent behaviors, and their properties in the frequency domain (Sec.~\ref{subsec_pm:motivation}).

Based on our analysis of these limitations, we propose a novel video demoiréing framework called the \textbf{Mo}iré-\textbf{C}onditioned \textbf{H}ybrid \textbf{A}daptive Transformer (MoCHA-former, Sec.~\ref{subsec_pm:overall}).
MoCHA-former comprises two key components: 1) \textbf{D}ecoupled \textbf{M}oiré \textbf{A}daptive \textbf{D}emoiréing (DMAD, Sec.~\ref{subsec_pm:dmad})  and 2) \textbf{S}patio-\textbf{T}emporal \textbf{A}daptive \textbf{D}emoiréing module (STAD, Sec.~\ref{subsec_pm:stad}). 
The DMAD module disentangles moiré patterns from content in the input moiréd frame, enabling adaptive demoiréing that accounts for the spatially varying characteristics of moiré artifacts. To this end, DMAD consists of three submodules: the \textbf{M}oiré-content \textbf{D}ecoupling \textbf{B}lock (MDB, Sec.~\ref{subsubsec_pm:MDB}), the \textbf{D}ecoupling \textbf{D}istillation \textbf{B}lock (DDB, Sec.~\ref{subsubsec_pm:ddb}), and the \textbf{M}oiré-adaptive \textbf{C}ross-attention \textbf{B}lock (MCB, Sec.~\ref{subsubsec_pm:MCB}). MDB separates moiré patterns and content using two convolution-based networks, producing moiré features and content features, respectively. DDB facilitates this separation by applying the Pre-Demoiré loss, Moiré-Cyclic loss, and Moiré Prediction loss, thereby enhancing the ability of MDB to effectively decouple moiré patterns from content. MCB then fuses the generated moiré and content features to produce moiré-adaptive features.
The STAD module addresses the large-scale characteristics of moiré patterns, the channel-dependent nature of the RAW domain, and the need for temporal consistency in video demoiréing. To achieve this, STAD incorporates \textbf{F}ourier \textbf{C}hannel \textbf{A}ttention (FCA, Sec.~\ref{subsubsec_pm:fca}) and a \textbf{W}indow-based \textbf{F}requency-adaptive Refinement \textbf{B}lock (WFB, Sec.~\ref{subsubsec_pm:wfb}) into a Video Swin Transformer \cite{vswin}. FCA is applied within the \textbf{S}patio-Temporal \textbf{F}requency-Aware Transformer \textbf{B}lock (SFB, Sec.~\ref{subsubsec_pm:sfb}) to consider the global properties of moiré patterns while adaptively aggregating inter-channel information in the Fourier domain by leveraging both spatio-temporal and frequency-domain cues. Inspired by the observation that moiré patterns exhibit distinct characteristics in the amplitude and phase components of the Fourier domain, WFB processes amplitude and phase separately. Unlike previous video demoiréing models~\cite{vdemoire,dtnet,rawvdemoire}, STAD performs frame alignment implicitly without using any explicit alignment module, based on our observation in Sec.~\ref{subsubsec_exp:alignment} that explicit alignment can adversely affect the demoiréing task. This implicit alignment strategy also enables STAD to handle abrupt temporal variations in moiré patterns.

Our contributions are summarized as follows:

\begin{itemize}
    \item We investigate the characteristics of moiré patterns by conducting experiments that analyze their impact on color distortion, spatially varying properties, channel-dependent behaviors, and significant temporal variations. 
    \item The proposed MoCHA-former adaptively removes moiré patterns while preserving fine details, and effectively processes spatio-temporal information to maintain both the global context of moiré patterns and temporal consistency across video frames.
    \item The proposed DMAD module separates moiré patterns from content through MDB and DDB, and then employs MCB to generate moiré-adaptive features that account for the spatially varying nature of moiré patterns.
    \item STAD leverages video window attention and shift operations to capture global information, while incorporating FCA and WFB in the frequency domain to exploit both the channel-dependent characteristics of moiré patterns and their frequency-domain properties for effective demoiréing.
    \item To validate the effectiveness of our approach, we conduct experiments on the RawVDemoiré and VDemoiré datasets. These experiments show that our model not only achieves state-of-the-art performance compared to previous video demoiréing models but also provides the highest parameter efficiency, as shown in Fig.~\ref{fig:psnr_lpips_param}.
\end{itemize}

%% file: sections/RelatedWorks.tex
\section{Related Works} \label{sec_rw:related_works}

\subsection{Image Demoiréing} \label{subsec_rw:image_demoireing}

Moiré patterns caused by frequency aliasing between the camera’s color filter array (CFA) and the display’s sub-pixel layout severely degrade image quality by obscuring content and introducing color distortions. To address this issue, earlier approaches rely on handcrafted prior-based models \cite{1,2}, which incorporate manually designed assumptions about moiré characteristics such as frequency distribution, texture patterns, or geometric structures. However, such models often fail to capture the wide range of moiré scales and intensities observed in real-world images.
With recent advances in deep learning and neural networks, learning-based methods \cite{vggnet, vit, transformer, alexnet} are now widely applied to the demoiréing task \cite{3,mopnet,awudn,14,19,20,fhde2net}, demonstrating significant improvements in handling complex moiré artifacts through architectural designs.
For example, \cite{3} introduces a multi-scale architecture to effectively handle moiré patterns of varying sizes, achieving notable performance gains.
In \cite{awudn,14,19,20}, a U-Net \cite{unet}-based encoder-decoder architecture is adopted to jointly leverage the detailed information in high-resolution features and the rich semantics in low-resolution features.
MopNet \cite{mopnet} proposes an attribute-aware classifier to simplify the representation of complex moiré patterns and applies channel-edge attention, inspired by the observation that moiré artifacts tend to concentrate around edge components.
FHDe²Net \cite{fhde2net} utilizes the frequency domain to separate moiré patterns from clean content, achieving strong performance.
Despite these advances, existing models \cite{1,2,3,awudn,14,19,20,mopnet,fhde2net} still suffer from two main limitations: (i) they fail to consider the spatially varying strength of moiré patterns within a single image, often leading to over-smoothing in image demoiréing, and they overlook the global characteristics of moiré patterns that affect the entire image. (ii) Moreover, applying image demoiréing models directly to video frames results in temporal inconsistencies, as these models are not designed to handle temporal information.
In this work, we address these challenges by proposing a video demoiréing model that explicitly separates moiré patterns and adaptively suppresses them based on their intensity. Our approach preserves fine details while effectively removing moiré patterns and jointly modeling spatio-temporal information, making it well-suited for video demoiréing tasks with better temporal consistency.

\subsection{Video Demoiréing} \label{subsec_rw:video_demoire}
While image demoiréing models have made significant progress, applying these methods to video in a frame-by-frame manner often results in a lack of temporal consistency \cite{vdemoire,rawvdemoire,dtnet}.
To address this issue, VDemoiré \cite{vdemoire} introduces a PCD alignment module based on EDVR \cite{edvr}, along with a multi-scale region-level relation loss. The PCD alignment module, in particular, has been adopted in several follow-up video demoiréing methods \cite{vdemoire, rawvdemoire, dtnet, fpanet}, demonstrating its effectiveness.
However, these methods introduce additional parameters for the alignment module and suffer from training instability \cite{basicvsr++, chan2021understanding}.
To maintain temporal consistency while considering long-range dependencies, STD-Net \cite{stdnet} alternates between spatial and temporal transformers. This design allows the model to capture global context and adaptively integrate temporal information. Nevertheless, it comes at the cost of significantly increased computational burden due to the large number of parameters required by vision transformer \cite{vit}.
To overcome these limitations, we propose Spatio-Temporal Adaptive Demoiréing (STAD), a lightweight and efficient module that reduces parameter overhead while effectively capturing both global context and spatio-temporal dependencies.

\subsection{sRGB and RAW Domain in Demoiréing} \label{subsec_rw:sgrb_raw_demoireing}
In general, images and videos captured by modern imaging devices are initially stored in the raw domain. These raw data are converted into sRGB format through a non-linear image signal processing (ISP) pipeline that includes operations such as white balancing, gamma correction, sharpening, and demosaicing \cite{isp}. When capturing digital displays, moiré patterns are introduced, and these artifacts become even more complex due to the non-linear characteristics of ISP operations (e.g., gamma correction, sharpening, demosaicing). As a result, moiré patterns spread across all color channels and become more difficult to remove \cite{tmm22, rrid, rawvdemoire}.
Early image and video demoiréing methods mainly rely on datasets in the sRGB domain \cite{lcdemoire, tip2018, fhde2net}. However, as discussed above, performing demoiréing in the sRGB domain is highly challenging due to the compounded effects of ISP operations.
To address this limitation, researchers have proposed datasets specifically in the raw domain, such as TMM22 dataset \cite{tmm22} for images  and RawVDemoiré \cite{rawvdemoire} for videos.
Based on these datasets, several raw-domain demoiréing methods \cite{tmm22,rrid,rawvdemoire} have been introduced.
RDNet \cite{tmm22} proposes a model that utilizes selective kernel attention to adaptively process moiré patterns, which exhibit different characteristics across color channels in the raw domain.
RRID \cite{rrid} leverages the relative simplicity of moiré removal in raw data while also incorporating sRGB data to perform joint color correction.
RawVDemoiré \cite{rawvdemoire} observes that moiré strength varies across channels in the raw domain, and introduces a method that applies learnable, channel-wise weighting to remove moiré patterns in an adaptive manner. We propose methods that leverage the channel-specific characteristics of the raw domain as well as the distinctive properties of moiré patterns in the raw domain.

\subsection{Degradation-Aware Restoration} \label{subsec_rw:degradation_aware}
The goal of restoration tasks in low-level vision is to remove degradations present in images or videos. Such restoration tasks can be mathematically formulated as follows:

\begin{equation}
\mathbf{y} \;=\; \mathcal{H}(\mathbf{x}) \;+\; \mathbf{n},    
\end{equation}
where $\mathbf{y}$ denotes the degraded image, $\mathbf{x}$ is the clean image, $\mathcal{H}(\cdot)$ represents the degradation operator, and $\mathbf{n}$ indicates noise. While most prior restoration methods \cite{restormer, basicvsr++, mopnet, rrid, 3, fhde2net,vdemoire,rawvdemoire,dtnet,stdnet} address this problem in an implicit manner, several works have proposed explicit degradation modeling \cite{fpro,dcpdn,dd_derain,tpami2023}. 
An implicit restoration approach is exemplified by Restormer \cite{restormer}, which performs restoration by learning a direct mapping from the input degraded image $\mathbf{y}$ to the ground-truth $\mathbf{x}$, without explicitly estimating the underlying degradation. Most representative demoiréing models \cite{mopnet, rrid, 3, fhde2net,vdemoire,rawvdemoire,dtnet,stdnet} follow a similar strategy, where moiré patterns are not directly predicted but instead removed through a direct mapping between moiréd inputs and clean targets.
In contrast, DCPDN \cite{dcpdn} adopts a physically-motivated dehazing model that explicitly estimates the transmission map and atmospheric light. JORDER \cite{dd_derain} explicitly separate and predict rain streaks and rain masks to handle the deraining task. FPro \cite{fpro} introduces a Dual Prompt Block (DPB) to represent degradations, leveraging separated high- and low-frequency information for targeted restoration. IDR \cite{idr} employs task-oriented knowledge collection to learn degradation-specific representations for image restoration. In the demoiréing domain, Wang et al. \cite{tpami2023} decouple moiré patterns and content and jointly optimize their concatenated features for restoration.
Motivated by these degradation-aware strategies, we design our model to explicitly decouple moiré patterns and content in the raw domain using dual branches and adaptive attention. This architecture guides the model to focus on moiré-affected regions, enabling effective moiré removal while preserving fine details.

%% file: sections/ProposedMethod.tex
\section{Proposed Method: MoCHA-former} \label{sec:proposed_method}
In this work, we propose a novel video demoiréing model that explicitly models the spatial–temporal characteristics of moiré patterns, their locally varying spatial periodicity and rapid frame‑to‑frame evolution, to remove artifacts while faithfully preserving fine details. To this end, we first analyze the fundamental properties of moiré patterns and video demoiréing datasets \cite{vdemoire,rawvdemoire}.

\subsection{Motivation and Key Insights} \label{subsec_pm:motivation}

\subsubsection{Color Distortion Caused by the Moiré Pattern.} \label{subsubsec_pm:color_distortion}

\begin{table}[t]
    \centering
    \begin{minipage}[t]{0.48\textwidth}
        \centering
        \begin{tabular}{l  c  c}
            \toprule
                         & \textbf{sRGB}   & \textbf{RAW} \\
            \midrule
            \textbf{Train} &  0.3468  &  0.4441 \\
            \textbf{Test}  &  0.3545  &  0.4877 \\
            \bottomrule
        \end{tabular}
        \caption{Average Color Correlation Between Moiré sRGB and Moiré RAW in the RawVDemoiré Dataset \cite{rawvdemoire} (color correlation of clean frame is 1). The color correlation values in the sRGB domain are noticeably lower compared to those in the RAW domain, indicating that color distortion is more severe in the sRGB domain.}
        \label{tab:color_corr}
    \end{minipage}
    \hfill
    \begin{minipage}[t]{0.48\textwidth}
        \centering
        \begin{tabular}{l  c  c}
            \toprule
                         & \textbf{Mean}   & \textbf{Variance} \\
            \midrule
            \textbf{Moiré} &  7389.37  &  7.1 $\times10^6$ \\
            \textbf{GT}    &  6378.11  &  2.2 $\times10^6$ \\
            \bottomrule
        \end{tabular}
        \caption{Quantitative result of amplitude differences between adjacent frames in the RawVDemoiré \cite{rawvdemoire} dataset. Moiréd videos exhibit greater temporal variation compared to clean videos.}
        \label{tab:temp_tab}
    \end{minipage}
\end{table}

\begin{figure}[t]
    \centerline{\includegraphics[width=\columnwidth]{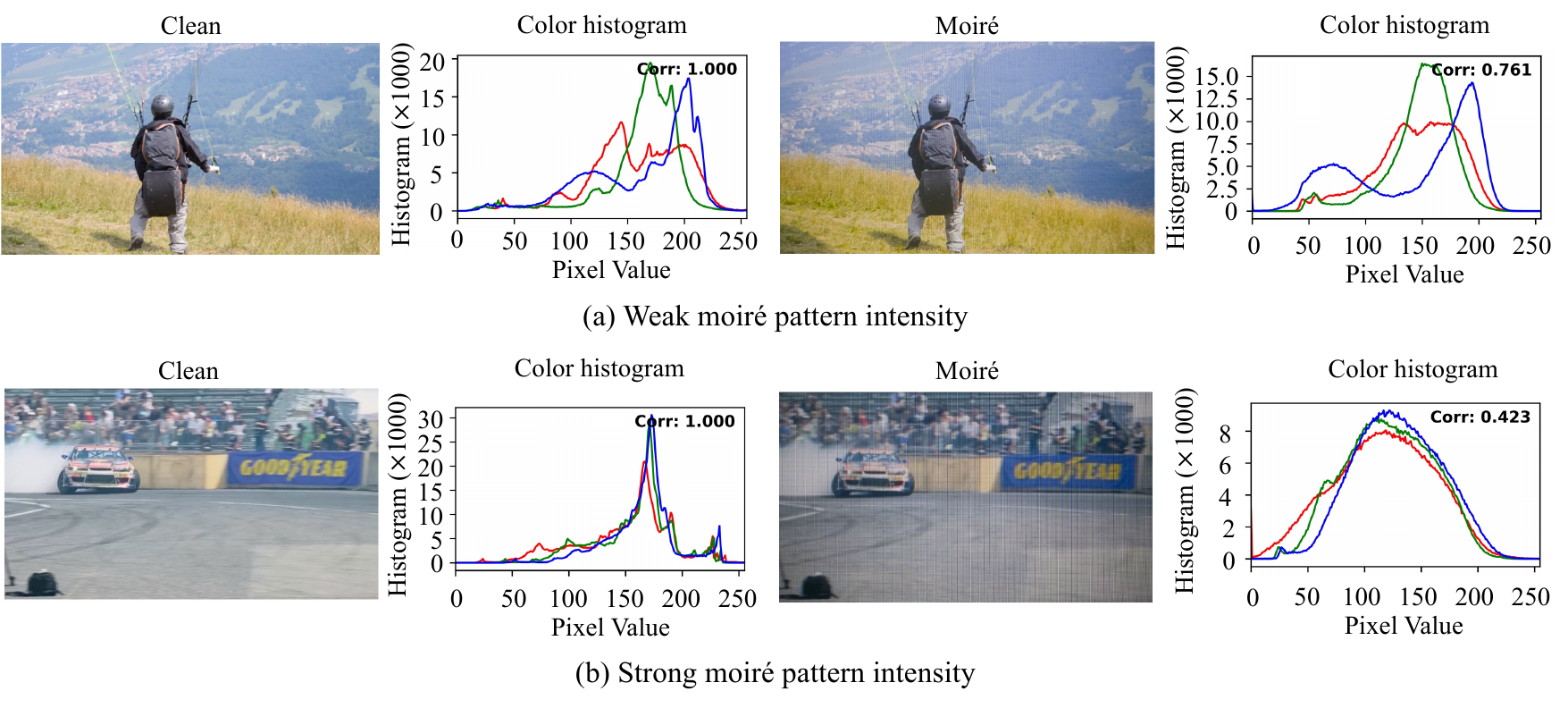}}
    \caption{
        Color histogram and color correlation comparison between clean and moiré images. The color histogram shows how much the color distribution shifts, while the color correlation quantitatively measures the degree of correlation drop, normalized such that the clean image has a value of 1. In (a), where the moiré pattern is weak, and (b), where it is strong, we observe that in the presence of stronger moiré patterns, i.e., in (b), both the shift in the color histogram and the drop in color correlation are more significant.}
    \label{fig:color_histo} 
\end{figure}

Moiré patterns not only obscure image content with their characteristic grid-like structures but also introduce significant color distortion \cite{fpanet,rawvdemoire,vdemoire}. This distortion alters the original colors of the content, leading to severe degradation of the visual quality. We analyze this color distortion both visually and quantitatively in Fig. \ref{fig:color_histo} and Table~\ref{tab:color_corr}. 
We quantify the inter-channel color correlation of an RGB frame $\mathbf{I} \in \mathbb{R}^{H \times W \times 3}$ by averaging the Pearson correlation coefficients between the three channel pairs $(R,G)$, $(G,B)$, and $(R,B)$. Formally, letting $|\Omega| = H \times W$ be the number of pixels and $I_{i,p}$ denote the value of channel $i \in \{R,G,B\}$ at pixel $p \in \Omega$, we define:

\begin{equation}
\mathrm{CC}(\mathbf{I}) \;=\; \frac{1}{3} \!
\sum_{(i,j)\in\{(R,G),(G,B),(R,B)\}}
\frac{\sum_{p \in \Omega} (I_{i,p} - \mu_i)(I_{j,p} - \mu_j)}
     {\sqrt{\sum_{p \in \Omega} (I_{i,p} - \mu_i)^2}\;
      \sqrt{\sum_{p \in \Omega} (I_{j,p} - \mu_j)^2}},
\label{eq:cc_def}
\end{equation}
where $\mu_i$ is the mean of channel $i$. To make the ground-truth (GT) image take the value $1$, we normalize by the GT correlation:
\begin{equation}
\widehat{\mathrm{CC}}(\mathbf{I})
\;=\;
\frac{\mathrm{CC}(\mathbf{I})}{\mathrm{CC}(\mathbf{I}_{\mathrm{GT}})},
\label{eq:cc_norm}
\end{equation}

so that $\widehat{\mathrm{CC}}(\mathbf{I}_{\mathrm{GT}})=1$. Consequently, the color correlation of a moiré-contaminated image $\mathbf{I}_{\mathrm{moire}}$ is given by
\begin{equation}
\widehat{\mathrm{CC}}(\mathbf{I}_{\mathrm{moire}})
\;=\;
\frac{\mathrm{CC}(\mathbf{I}_{\mathrm{moire}})}
     {\mathrm{CC}(\mathbf{I}_{\mathrm{GT}})}.
\label{eq:cc_moire}
\end{equation}
Values of $\widehat{\mathrm{CC}}$ close to $1$ indicate that the inter-channel statistical structure is well preserved relative to the GT, whereas lower values reflect a collapse or distortion of channel correlations induced by moiré artifacts.
Fig. \ref{fig:color_histo}-(a) and (b) show how the color histogram decreases as the severity of the moiré pattern increases. A lower color correlation indicates stronger color distortion. Additionally, as shown in Table~\ref{tab:color_corr}, the color correlation in the sRGB domain is consistently lower than that in the RAW domain. This result demonstrates that color distortion is more pronounced in the sRGB domain compared to the RAW domain. As shown in the color histograms in Fig. \ref{fig:color_histo}-(a) and (b), the patterns of color distortion vary across channels. This channel-wise inconsistency introduces additional challenges in removing color distortion from moiré-contaminated frames. To address the issue of channel-wise color distortion, we introduce frequency-based channel attention mechanisms in both the FCA (Sec.~\ref{subsubsec_pm:fca}) and the WFB (Sec.~\ref{subsubsec_pm:wfb}). These modules enable adaptive processing of important information through enhanced channel interaction in the frequency domain.

\begin{figure}[!t]
    \centerline{\includegraphics[width=\columnwidth]{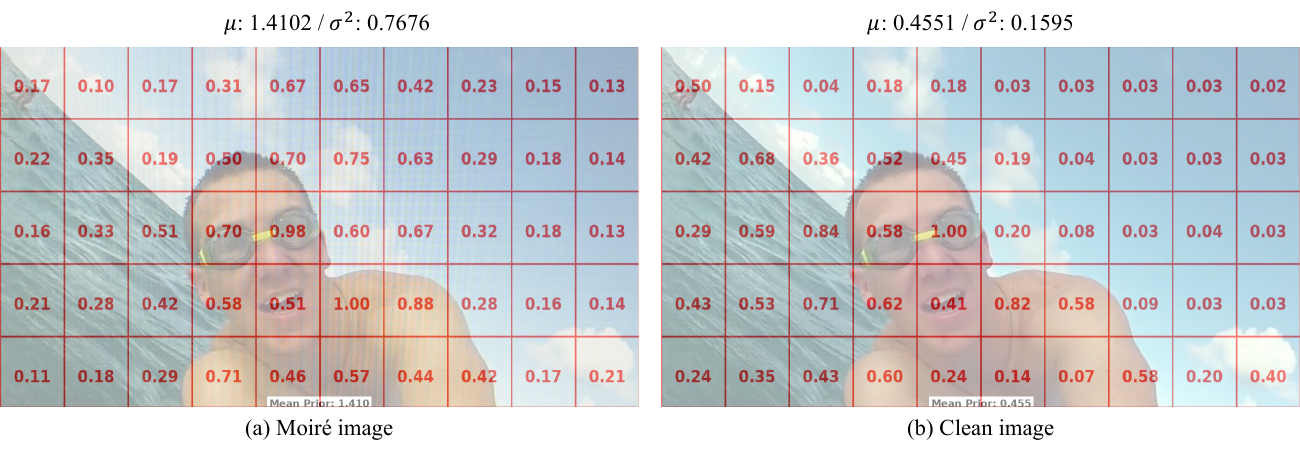}}
    \caption{Qualitative result for inter-patch complexity based on the DDA moiré prior \cite{dda}. The DDA framework \cite{dda} introduces a moiré prior that quantitatively measures the levels of colorfulness and frequency. Compared to the clean image (b), the moiré image (a) exhibits both higher mean and variance. This indicates that the moiré prior values of individual patches are higher in the moiré image, and the variation of moiré prior values across patches is also greater.}
    \label{fig:iclr_dda} 
\end{figure}

\subsubsection{Spatially Varying Intensity Demonstrated in a Single Image by the Moiré Pattern}. \label{subsubsec_pm:moire_single_image}

\begin{table}[t]
\centering
\begin{tabular}{lcc}
\toprule
 & \multicolumn{1}{c}{\textbf{Clean}} & \multicolumn{1}{c}{\textbf{Moiré}} \\
 & \textbf{Mean} / \textbf{Var.} & \textbf{Mean} / \textbf{Var.} \\
\midrule
\textbf{RawVDemoire} \cite{rawvdemoire}    & 0.5782 / 0.0761 & 0.7737 / 0.3217 \\
\textbf{VDemoire}    \cite{vdemoire}       & 0.4478 / 0.0390 & 0.5041 / 0.0414 \\
\bottomrule
\end{tabular}
\caption{Quantitative analysis of moiré prior values based on DDA \cite{dda}. In both the RawVDemoiré \cite{rawvdemoire} and VDemoiré \cite{vdemoire} datasets, moiré-contaminated frames exhibit higher mean and variance of moiré prior values compared to clean frames. This indicates that the intra-frame patch-wise complexity is greater in moiré frames.}
\label{tab:iclr_dda}
\end{table}

The intensity of moiré patterns varies even within a single image \cite{tip2018, mopnet, moirepose, dda}. To support this observation, DDA \cite{dda} introduces the concept of a moiré prior, which quantifies the intensity of colorfulness and frequency of moiré patterns. Following this approach, we divide each input image into $128\times128$ patches and compute the moiré prior for each patch. The results are presented in Fig. \ref{fig:iclr_dda} and Table~\ref{tab:iclr_dda}. As shown in Fig. \ref{fig:iclr_dda}, when comparing the mean and variance of the summed moiré prior values across patches between clean images and moiré-contaminated images, both the mean and variance are higher in moiré images. This indicates that moiré images exhibit greater overall diversity in terms of colorfulness and frequency compared to clean images. At the same time, it suggests that the variations in colorfulness and frequency across different patches are also larger in moiré images than in clean images. This observation supports the fact that moiré-contaminated images exhibit varying levels of intensity and complexity even within the single image. This tendency is consistently observed in both the RawVDemoiré \cite{rawvdemoire} and VDemoiré \cite{vdemoire} datasets, as summarized in Table~\ref{tab:iclr_dda}.
To adaptively capture such intra-image variations in moiré patterns and deliver more informative features to subsequent modules, we propose the Decoupled Moiré-Adaptive Demoiréing (DMAD, Sec.~\ref{subsec_pm:dmad}) module.

\subsubsection{Temporal Characteristics of Moiré Patterns.} \label{subsubsec_pm:temporal}

\begin{figure}[!t]
    \centerline{\includegraphics[width=\columnwidth]{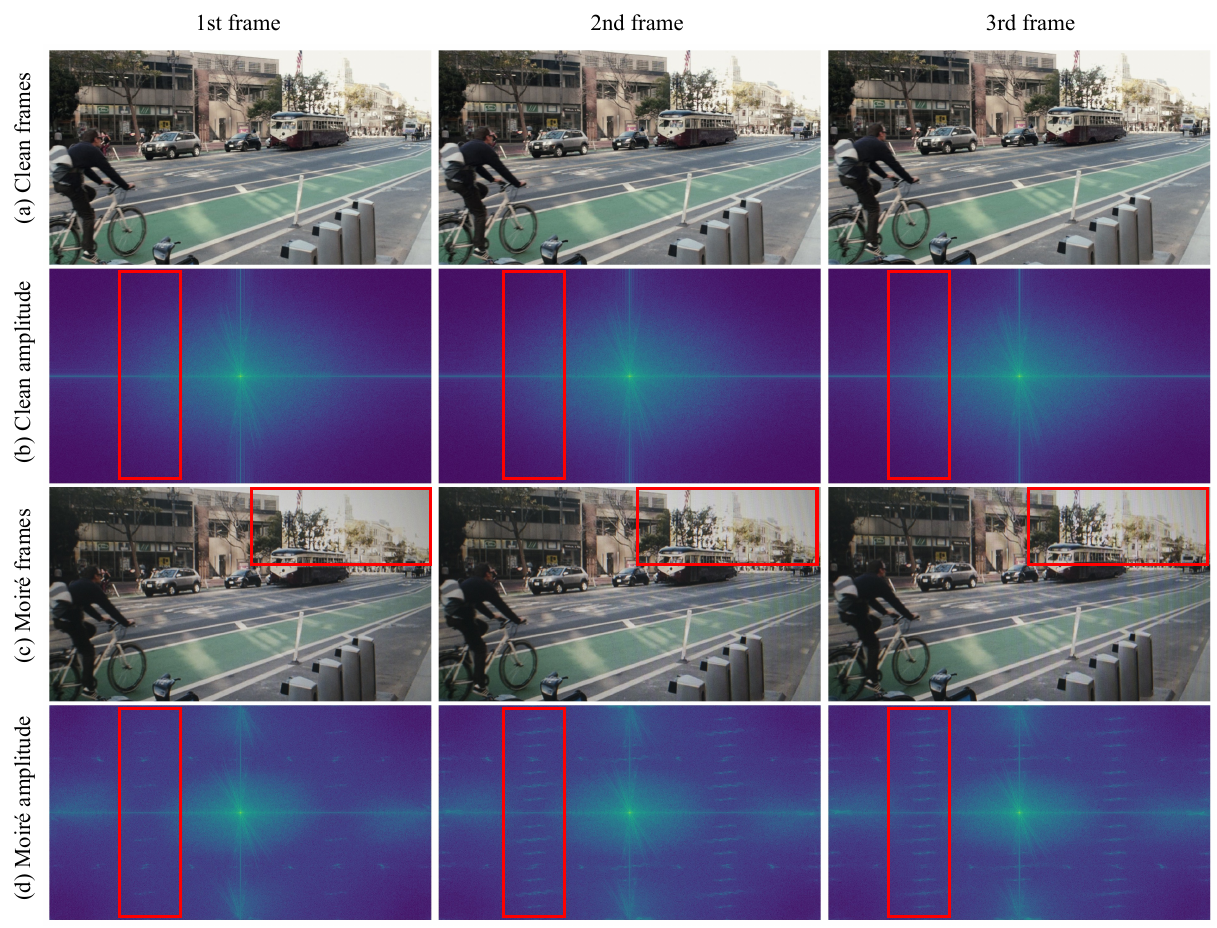}}
    \caption{
        A qualitative result demonstrating significant moiré pattern variation across consecutive frames in the moiré video dataset~\cite{rawvdemoire}. By comparing the amplitude of each frame, we observe that the temporal variation in amplitude is more pronounced in (d) than in (b). As observed in the red box of row (c), the characteristics of the moiré patterns vary significantly over time, which is further evidenced by the substantial changes in the patterns highlighted in the red box of row (d).}
    \label{fig:moire_temp_vis} 
\end{figure}

Moiré patterns are caused by frequency aliasing between the screen’s sub-pixels and the camera’s color filter array (CFA), and their appearance changes significantly as the camera’s position and orientation vary \cite{mopnet,rrid, esdnet, tpami2023, tmm22, awudn}. The VDemoiré dataset \cite{vdemoire} and RawVDemoiré dataset \cite{rawvdemoire} consist of videos captured in a hand-held manner. Due to the nature of hand-held acquisition, subtle movements of the recording device are inevitable, and even minor motions can lead to significant variations in the appearance of moiré patterns, given their frequency aliasing origin. As shown in Fig. \ref{fig:moire_temp_vis}, we extract the amplitude from three consecutive frames of a clean video clip and its corresponding moiré-contaminated video clip. Notably, the temporal variations in amplitude across moiré frames (fourth row of Fig. \ref{fig:moire_temp_vis}) are much more pronounced than those observed in clean frames (second row of Fig. \ref{fig:moire_temp_vis}). Table~\ref{tab:temp_tab} presents the mean and variance of amplitude differences between adjacent frames in the RAW video dataset \cite{rawvdemoire}. The amplitudes are extracted by applying FFT to each frame, and the absolute differences between adjacent frames are summed to compute the temporal frequency variation. A larger value indicates greater temporal variation in the frequency domain. As shown in Table~\ref{tab:temp_tab}, moiré-contaminated frames exhibit both higher mean and variance compared to clean frames. These observations highlight that moiré patterns exhibit highly dynamic variations over time. To address this, we propose a Spatio-temporal Frequency-aware Transformer Block (SFB, Sec.~\ref{subsubsec_pm:sfb}) that adaptively captures such temporal variations while jointly leveraging spatial and frequency information.

\subsubsection{Differences Between sRGB and RAW Domains in Moiré Patterns.} \label{subsubsec_pm:raw_rgb}

\begin{figure}[!t]
    \centerline{\includegraphics[width=\columnwidth]{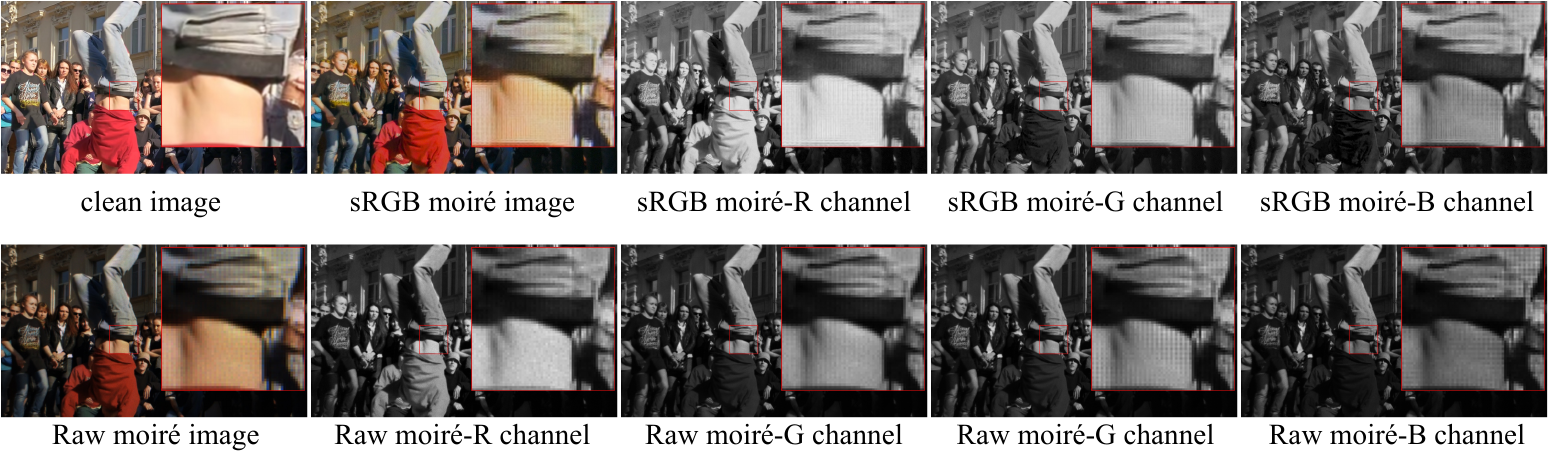}}
    \caption{Qualitative result of channel-wise moiré pattern characteristics in sRGB and RAW images. In sRGB images (top row), the moiré pattern is evenly distributed across all channels, whereas in RAW images (bottom row), the moiré pattern exhibits distinct characteristics for each channel. We recommend zooming in for more detail.}
    \label{fig:rgb_raw_ch_moire} 
\end{figure}

Moiré patterns exhibit different characteristics in the sRGB and RAW domains \cite{rawvdemoire,rrid,tmm22}. In the RAW domain, moiré patterns appear with varying intensities and patterns across different channels, whereas in the sRGB domain, they tend to exhibit similar intensities across all channels regardless of channel type. This difference arises from the application of the image signal processing (ISP) pipeline, which involves non-linear operations during the conversion from RAW to sRGB. As a result, moiré patterns are propagated across all channels in the sRGB domain \cite{rawvdemoire}. As shown in Fig. \ref{fig:rgb_raw_ch_moire}, moiré patterns appear strongly in all R, G, and B channels of the sRGB image. In contrast, in the RAW frame, moiré patterns are prominently visible in the two G channels, while they appear relatively weaker in the R and B channels. To address this channel-dependent behavior of moiré patterns, we introduce a Fourier Channel Attention (FCA, Sec.~\ref{subsubsec_pm:fca}) mechanism that explicitly accounts for these differences in our proposed model.

\subsubsection{Frequency Domain Characteristics of Moiré Patterns.} \label{subsubsec_pm:frequency}

\begin{figure}[!t]
    \centerline{\includegraphics[width=\columnwidth]{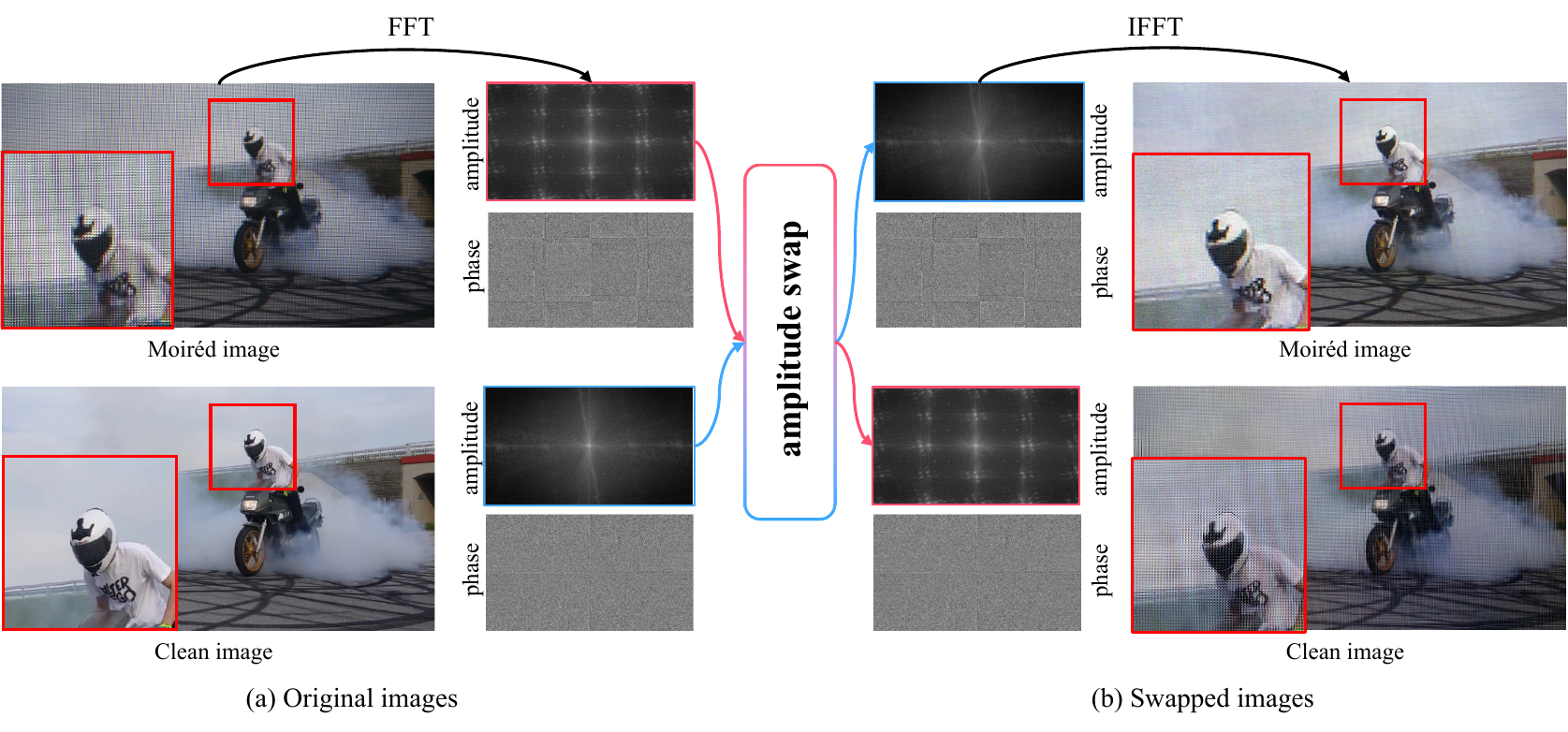}}
    \caption{Qualitative result showing the distinct information encoded in the amplitude and phase components of the FFT. In (a), FFT is applied to both the moiré image and the clean image to extract their amplitude and phase. In (b), the amplitude components are swapped. The upper part of (b) reveals the appearance of the moiré pattern, while the lower part highlights its structural characteristics.}
    \label{fig:fft_amp_pha_swap} 
\end{figure}

Several methods \cite{fhde2net, dmcnn, awudn, tmm22, fpanet} utilize the frequency domain to address the characteristics of moiré patterns. We observe that the amplitude component of a moiré image contains structural characteristics of moiré patterns (e.g., grid-like structures), while the phase component captures appearance-level attributes (e.g., smoothing of the original content). As shown in Fig.~\ref{fig:fft_amp_pha_swap}, we extract amplitude and phase by applying FFT to both a moiré-contaminated image and a clean image, and then swap the amplitude components between the two before reconstructing the images via inverse fast fourier transform (IFFT).
As shown in Fig.~\ref{fig:fft_amp_pha_swap}-(b), the upper image, which combines the phase of the moiré image and the amplitude of the clean image, reveals appearance-level characteristics of the moiré pattern, such as smoothing artifacts and color distortion. In contrast, the lower image, which combines the amplitude of the moiré frame and the phase of the clean frame, highlights structural aspects of the moiré pattern, including grid-like formations. These observations suggest that structural information unique to moiré patterns, such as grid textures, is primarily encoded in the amplitude component, while appearance-level effects like smoothing and color distortion are predominantly captured in the phase component.
These observations highlight the benefit of processing amplitude and phase separately when working in the frequency domain for moiré pattern removal. Motivated by this, we propose a Window-based Frequency Adaptive Refinement Block (WFB, Sec.~\ref{subsubsec_pm:wfb}) that explicitly separates and processes the amplitude and phase components.

\begin{figure}[!t]
    \centerline{\includegraphics[width=\columnwidth]{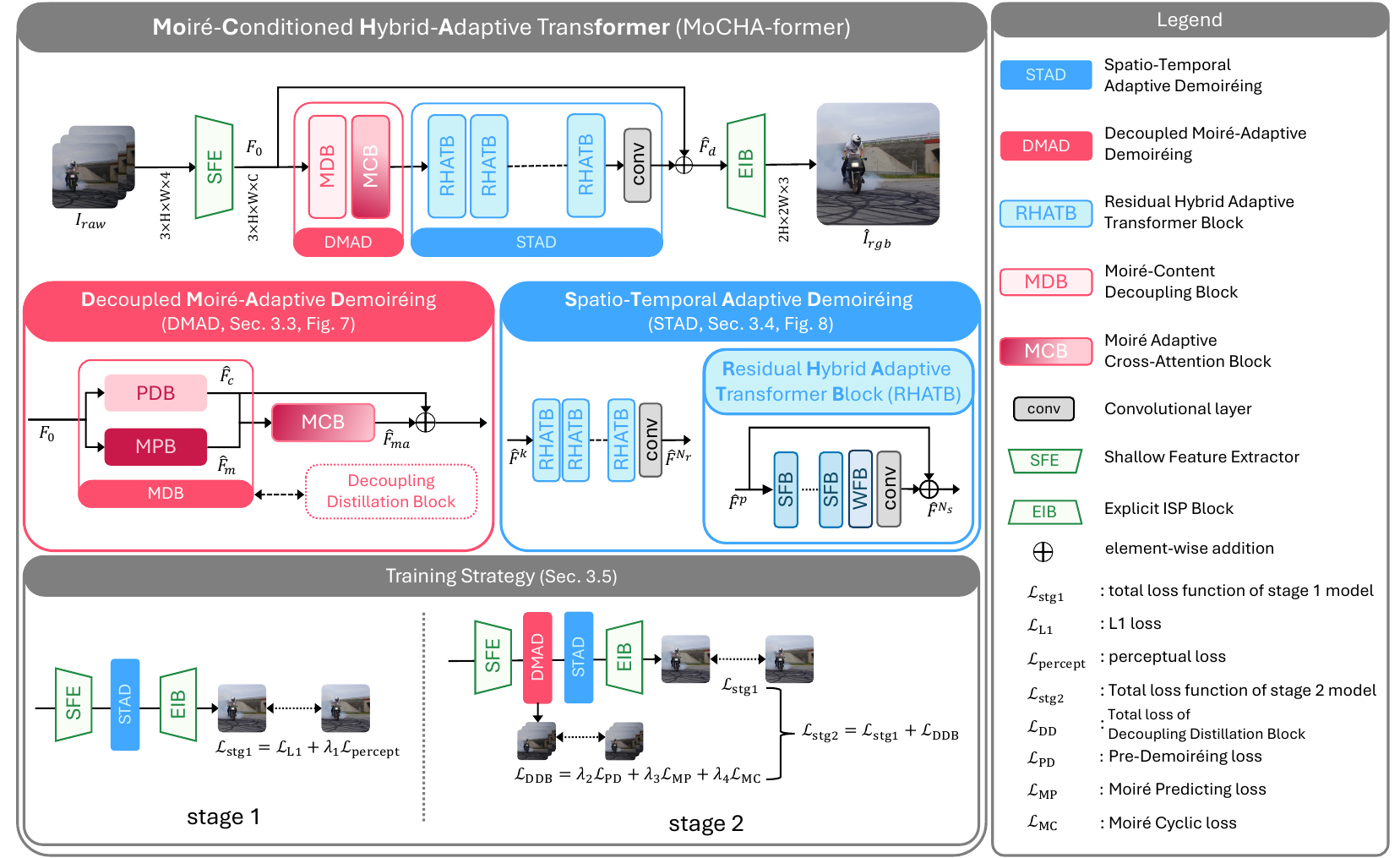}}
    \caption{Overview of our proposed video demoiréing framework, Moire-Condtioned Hybrid-Adaptive Transformer (MoCHA-former). The overall framework consists two key modules. (1) Decoupled Moiré-Adaptive Demoiréing module (DMAD) and (2) Spatio-Temporal Adaptive Demoiréing module (STAD).}
    \label{fig:framework_overview} 
\end{figure}

\subsection{Overall Framework} \label{subsec_pm:overall}

In this section, we introduce novel framework, MoCHA-former, composed of DMAD (Decoupled Moiré‑Adaptive Demoiréing) and STAD (Spatio‑Temporal Adaptive Demoiréing), which separates moiré and content in the raw domain and then integrates spatio–temporal and frequency cues for robust, detail‑preserving video demoiréing as shown in Fig.~\ref{fig:framework_overview}. MoCHA-former is composed of two stages. First, we present the Decoupled Moiré-Adaptive Demoiréing (DMAD) module (Sec. \ref{subsec_pm:dmad}), which adaptively removes diverse moiré patterns that appear even within a single frame. Second, we introduce the Spatio-Temporal Adaptive Demoiréing (STAD) module (Sec. \ref{subsec_pm:stad}), which simultaneously leverages spatio-temporal information and frequency domain representations for moiré pattern removal.

\textbf{Overall Framework.} Our framework takes three consecutive RAW frames, denoted as $\mathbf{I_{raw}}=\{I_{raw}^{t-1},I_{raw}^{t},I_{raw}^{t+1}\}$, as input and outputs a single demoiréd center-time index of $t$ sRGB frame $\hat{I}_{rgb}$. Each input frame $I_{raw}^{t-1},I_{raw}^{t}I_{raw}^{t+1} \in \mathbb{R}^{H\times W\times 4}$ is a RAW domain image with four channels, RGGB, while the output frame $\hat{I}_{rgb} \in \mathbb{R}^{2H\times 2W\times 3}$ is an sRGB frame with spatial resolution doubled compared to the input.
The consecutive input frames $\mathbf{I_{raw}}$ are first processed by a Shallow Feature Extractor (SFE), consisting of a single $3\times3$ convolutional layer ($3\times3 conv$ ), to obtain the moiré-corrupted features $F_{0} \in \mathbb{R}^{3\times H\times W\times C}$. These features are then fed into the DMAD module to generate moiré-adaptive features $\hat{F}_{ma}$, which adaptively account for the diverse moiré patterns present in the input.
Next, $\hat{F}_{ma}$ is passed to the STAD module, which jointly exploits spatio-temporal information and frequency-domain representations to produce the demoiréd features $\hat{F}_d$. These demoiréd features are combined with the original moiré-corrupted features $F_0$ through a skip connection \cite{swinir} and forwarded to the next stage.
Finally, the combined features are processed by the Explicit ISP Block (EIB), consisting of two convolutional layers followed by a pixel-shuffle layer, to reconstruct the final single sRGB output frame $\hat{I}_{rgb}$.

\subsection{Decoupled Moiré Adaptive Demoiréing module (DMAD)} \label{subsec_pm:dmad}

\begin{figure}[!t]
    \centerline{\includegraphics[width=0.75\columnwidth]{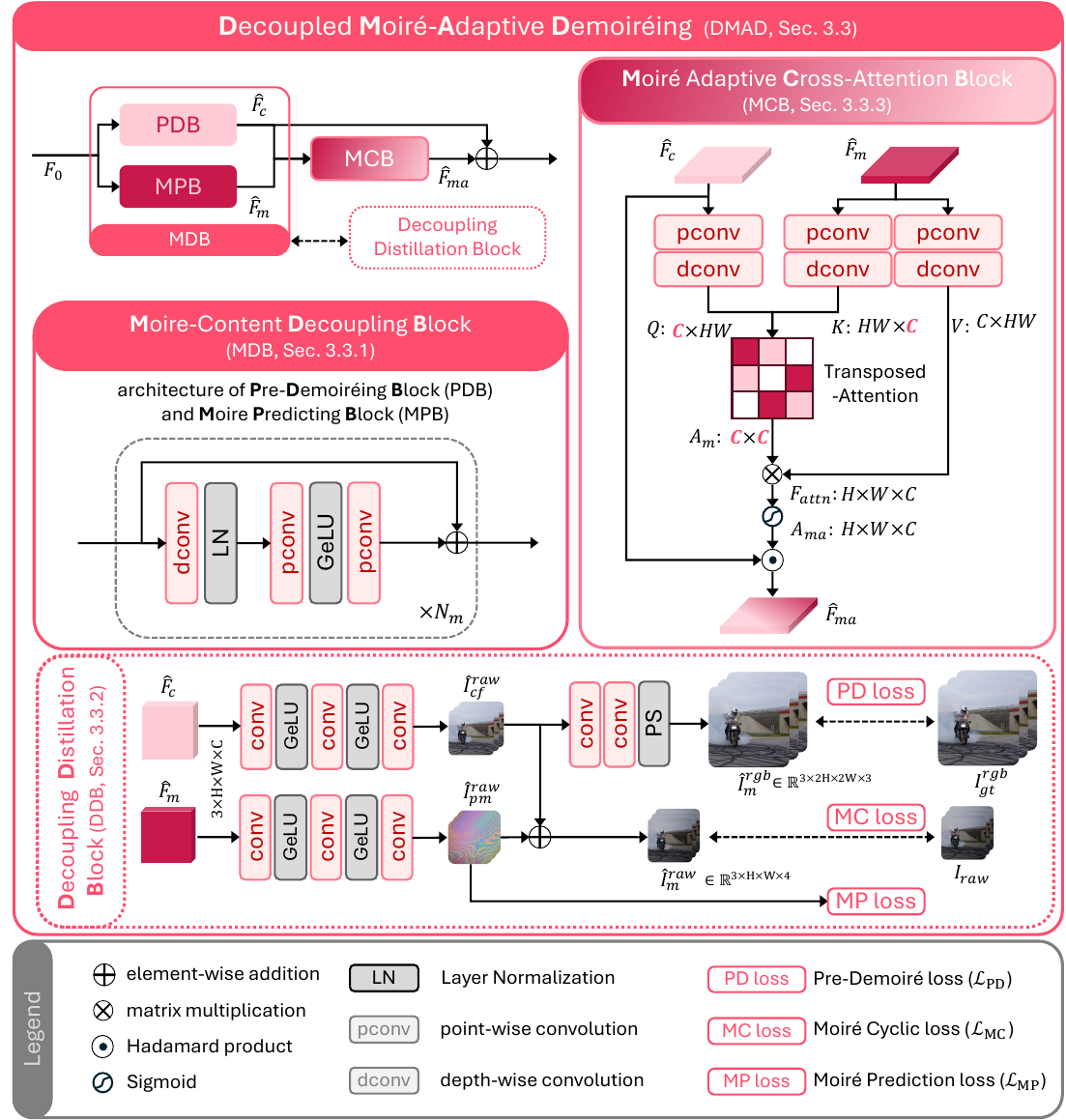}}
    \caption{Overview of our proposed Decoupled Moiré-Adaptive Demoiréing (DMAD, Sec.~\ref{subsec_pm:dmad}). The DMAD consists of the Moiré-Content Decoupling Block (MDB, Sec.~\ref{subsubsec_pm:MDB}), which separates moiré patterns from content, the Decoupling Distillation Block (DDB, Sec.~\ref{subsubsec_pm:ddb}) that facilitates the decoupling performed by the MDB, and the Moiré Adaptive Cross-Attention Block (MCB, Sec.~\ref{subsubsec_pm:MCB}), which leverages the decoupled features to generate moiré-adaptive representations.}
    \label{fig:dmad} 
\end{figure}

As shown in Fig. \ref{fig:dmad}, the DMAD consists of two main modules: 1) the \textbf{M}oire-Content \textbf{D}ecoupling \textbf{B}lock (MDB), and 2) the \textbf{M}oiré-Adaptive \textbf{C}ross-Attention \textbf{B}lock (MCB). The goal of the MDB is to decouple the moiré pattern from the content and generate moiré-adaptive features using the predicted moiré pattern as guidance. In the MCB, both the predicted moiré pattern and the clear content obtained from the MDB are utilized to generate features that are adaptively conditioned on the moiré pattern. In summary, the DMAD enables the model to focus adaptively on regions affected by moiré patterns and refines the features to effectively remove moiré artifacts while preserving fine details.

\subsubsection{Moiré-Content Decoupling Block (MDB)} \label{subsubsec_pm:MDB}
To effectively decouple moiré patterns from content, the input feature $F_0$ is separately fed into the \textbf{P}re-\textbf{D}emoiréing \textbf{B}lock (PDB) and the \textbf{M}oire \textbf{P}redicting \textbf{B}lock (MPB). Both PDB and MPB adopt convolution-based architectures to preserve local details and high-frequency components of the moiré pattern, following the findings in \cite{how_vit}. As shown in Fig.~\ref{fig:dmad}, both the PDB and MPB adopt convolution-based models with $N_m$ layers, respectively. Using the shallow feature $F_0$ as input, the PDB outputs the partially demoiréd feature $\hat{F}_c$, while the MPB predicts the moiré-specific feature $\hat{F}_m$, which contains only the moiré pattern components. These two features, $\hat{F}_c$ and $\hat{F}_m$, are obtained as follows:

\begin{equation}
{F}_{out} = {F}_{in} + \sum_{i=1}^{N_m} \mathcal{B}_i(F_{in}),
\label{eq:mdb_residual}
\end{equation}

\begin{equation}
\mathcal{B}(F) = pconv_2\left( GeLU\left( pconv_1\left( LN\left( dconv(F) \right) \right) \right) \right),
\label{eq:mdb_block}
\end{equation}
$pconv$ denotes a $1 \times 1$ point-wise convolution, $GeLU$ refers to the activation function proposed in \cite{gelu}, $LN$ indicates layer normalization \cite{ln}, and $dconv$ represents depth-wise convolution. Simply splitting the network into two branches does not guarantee effective separation of moiré patterns and content. To support this decoupling process, we propose the \textbf{D}ecoupling \textbf{D}istillation \textbf{B}lock (DDB), which is detailed in Sec. \ref{subsubsec_pm:ddb}.

\subsubsection{Decoupling Distillation Block (DDB)} \label{subsubsec_pm:ddb}

As previously mentioned in Sec. \ref{subsubsec_pm:MDB}, the Decoupling Distillation Block (DDB) imposes constraints to facilitate the separation of moiré patterns and content. This block distills useful knowledge by processing both the GT frame and the moiré-contaminated input through a series of convolutional layers, thereby guiding the PDB and MPB to more effectively decouple content from moiré patterns. 

The DDB takes as input the pre-demoiréd feature $\hat{F}_c$ and the predicted moiré feature $\hat{F}_m$. Each feature is processed through several $3\times3$ convolutional layers and GeLU activate function to generate the predicted demoiréd frames $\hat{I}^{raw}_{cf} \in \mathbb{R}^{T \times H \times W \times 4}$ and the predicted moiré pattern frames $\hat{I}^{raw}_{pm} \in \mathbb{R}^{T \times H \times W \times 4}$.

Since the GT RAW frames provided in RawVDemoiré \cite{rawvdemoire} are pseudo GTs obtained by converting GT sRGB frames, directly using these frames can lead to inaccurate supervision. Therefore, $\hat{I}^{raw}_{cf}$ is further processed using two $3\times3$ convolutional layers followed by a pixel-shuffle layer to produce sRGB frames $\hat{I}^{rgb}_{m}$. To enhance demoiréing performance, we apply a \textbf{P}re-\textbf{D}emoiréing loss (PD loss) as:

\begin{equation}
\mathcal{L}_{\mathrm{PD}}
= \frac{1}{|THW|} \Bigl\| \\ \hat{I}^{rgb}_{m} - {I}^{rgb}_{gt} \Bigr\|_{2},
\label{eq:pd_loss}
\end{equation}
$T$, $H$, and $W$ denote the number of frames, the height, and the width of each frame, respectively. $\hat{I}^{rgb}_{m}$ represents the predicted demoiréd RGB frames, while ${I}^{rgb}_{gt}$ indicates the corresponding ground-truth (GT) RGB frames. The PD loss is computed as the $L_2$ loss between frames. This loss supervises the network to remove moiré patterns effectively.

For $\hat{I}^{raw}_{pm}$, considering that moiré patterns inherently exhibit self-similarity, we apply a \textbf{M}oiré \textbf{P}rediction loss (MP loss) to encourage the separation of moiré patterns from content. This loss is designed to enhance self-similarity based on Weighted Nuclear Norm Minimization (WNNM) \cite{wnnm} as:

\begin{equation}
\mathcal{L}_{\mathrm{MP}}
= \sum_{g=1}^{G} \omega_{g}\, \bigl\| \Sigma_s\!\bigl( \mathbf{P}_{g}(\hat{I}^{\mathrm{raw}}_{pm}) \bigr) \bigr\|_{*}.
\label{eq:mp_loss}
\end{equation}

The Moiré Prediction loss $\mathcal{L}_{\mathrm{MP}}$ encourages the predicted moiré‑only frame
$\hat{\mathbf{I}}^{\mathrm{raw}}_{pm}$ to exhibit strong internal self‑similarity,  
a characteristic of real moiré patterns.  
For each of the $G$ patch groups found by non‑local similarity search,  
$\mathbf{P}_{g}(\hat{I}^{\mathrm{raw}}_{pm})$ stacks all patches in the $g$‑th group as column vectors,  
forming a matrix whose singular values are collected by $\Sigma_s(\,\cdot\,)$.
The nuclear norm $\lVert\cdot\rVert_{*}$ (the $\ell_{1}$ norm of these singular values) penalizes rank,  
promoting redundancy within each group. The scalars $\omega_{g}$ are adaptively determined via Weighted Nuclear Norm Minimization (WNNM). They assign larger penalties to smaller singular values so that dominant self‑similar structures are preserved. Adding over all groups yields a global measure that guides the network to isolate coherent moiré patterns.

However, applying only this loss can lead to trivial solutions, as noted in \cite{tpami2023}. To stabilize training and avoid trivial solutions, we incorporate a \textbf{M}oiré \textbf{C}yclic loss (MC loss). Specifically, we sum the predicted demoiréd frames $\hat{I}^{raw}_{cf}$ and the predicted moiré pattern $\hat{I}^{raw}_{pm}$ and supervise them (using L2 loss) against the input moiré frames $I_{raw}$:

\begin{equation}
\mathcal{L}_{\mathrm{MC}}
= \frac{1}{|THW|} \Bigl\|(\hat{I}^{raw}_{cf} + \hat{I}^{raw}_{pm}) - {I}^{raw} \Bigr\|_{2}.
\label{eq:mc_loss}
\end{equation}

\begin{figure}[!t]
    \centerline{\includegraphics[width=0.5\columnwidth]{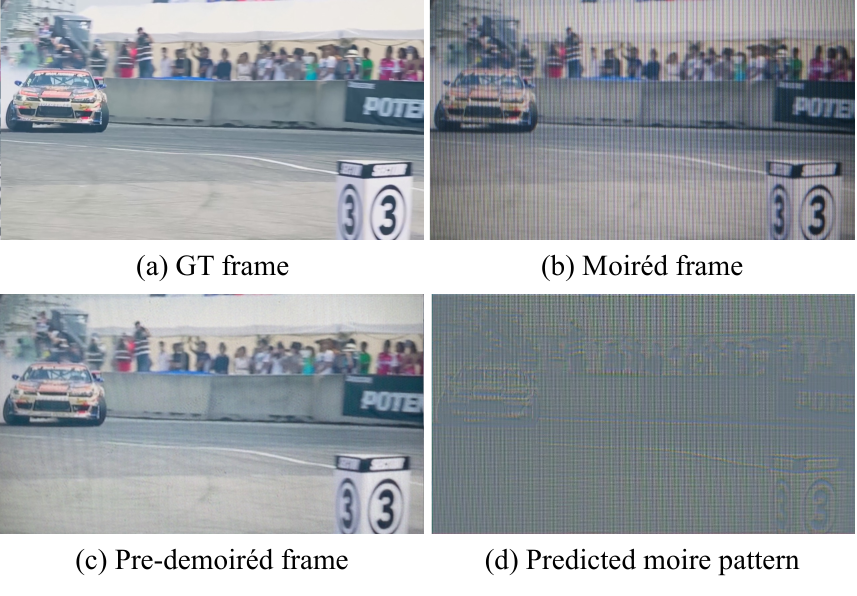}}
    \caption{Qualitative result of our proposed MDB (Sec.~\ref{subsubsec_pm:MDB}) and DDB (Sec.~\ref{subsubsec_pm:ddb}) module. (a) represents the ground-truth frame, (b) shows the corresponding moiréd frame, (c) depicts the pre-demoiréd frame generated by the PDB, and (d) illustrates the predicted moiré pattern produced by the MPB. As observed in (c) and (d), the moiré patterns and content are effectively disentangled.}
    \label{fig:dmad_vis} 
\end{figure}

Through this process, the MDB effectively separates moiré patterns from content and passes the decoupled features to the subsequent module.
Ultimately, the DDB serves as a dedicated module for optimizing the DMAD, and the losses described above are integrated into the overall training objective as defined in Eq.~\ref{eq:total_loss}. The qualitative results of DMAD with the inclusion of DDB are presented in Fig.~\ref{fig:dmad_vis}. As shown in Fig.~\ref{fig:dmad_vis}, the proposed DMAD effectively separates moiré patterns from content in moiréd frames.

\subsubsection{Moire Adaptive Cross-Attention Block (MCB)} \label{subsubsec_pm:MCB}
To perform moiré‑adaptive demoiréing, we propose the \textbf{M}oiré‑Adaptive \textbf{C}ross‑\textbf{A}ttention Block (MCB), which fuses the pre‑demoiréd feature $\hat{F}_c$ and the predicted moiré feature $\hat{F}_m$, both of which have been effectively decoupled by the DDB described in Sec.~\ref{subsubsec_pm:ddb}. The MCB leverages cross-attention to adaptively fuse the predicted clear content and moiré pattern features. This mechanism enables the model to effectively explore the locally varying characteristics of moiré patterns across different spatial regions. Firstly, the two predicted features are embedded to achieve above effects. To capture spatial information, we apply depth-wise convolution, followed by a $1\times1$ point-wise convolution to enable channel-wise interaction, producing the query $Q \in \mathbb{R}^{C \times HW}$, key $K \in \mathbb{R}^{HW \times C}$, and value $V \in \mathbb{R}^{C \times HW}$ for the attention computation:

\begin{equation}
Q = {dconv}({pconv}(\hat{F}_c)), \quad 
\{K, V\} = {dconv}({pconv}(\hat{F}_m)).
\end{equation}

Here, $dconv$ denotes a depth-wise convolution layer, and $pconv$ denotes a $1\times1$ point-wise convolution layer. We adopt transposed cross-attention \cite{restormer} instead of standard attention \cite{vit}. There are two main reasons for this design choice. First, transposed cross-attention allows us to model the correlation between the two features in a global context. As discussed in Sec. \ref{subsubsec_pm:moire_single_image}, moiré patterns often spread across wide regions within a single frame and exhibit spatially varying intensity and structure. From this perspective, a global interaction is essential, and cross-attention is well-suited to dynamically compute the correlation between content and moiré patterns across the entire frame.
Second, considering the characteristics of RAW domain frames and aiming to effectively remove color distortion, it is crucial to enable information exchange across channels. As noted in Sec. \ref{subsubsec_pm:raw_rgb}, moiré patterns in RAW domain frames exhibit different characteristics across channels. Furthermore, as discussed in Sec. \ref{subsubsec_pm:color_distortion}, color distortion introduced by moiré patterns also varies channel-wise. To account for these discrepancies, we first embed the input through convolutional layers that consider channel-wise differences. Then, we apply transposed attention to facilitate channel-level interactions. This makes transposed cross-attention a well-justified and effective choice for handling the unique characteristics of moiré patterns in RAW domain inputs.
Based on the above observations, we introduce the transposed cross-attention operation to construct a moiré-adaptive attention map:

\begin{equation}
A_{m} = \mathrm{Softmax}(Q \cdot K / \lambda).
\end{equation}

After generating the transposed attention map $A_{m}$, we apply the attention operation to compute the feature map as:

$\hat{F}_{attn}$ as follows:
\begin{equation}
\hat{F}_{attn} = V \cdot A_m.
\end{equation}

Next, we apply a sigmoid activation to obtain the moiré-adaptive attention map $A_{ma}$:

\begin{equation}
A_{ma} = \mathrm{Sigmoid}(\hat{F}_{attn}).
\end{equation}

The resulting attention map $A_{ma}$ can be interpreted as an attention map that assigns higher scores to regions strongly affected by moiré patterns. We then perform a Hadamard product between the pre-demoiréd feature $\hat{F}_c$ and the moiré-adaptive attention map $A_{ma}$ to generate the moiré-adaptive feature map $\hat{F}_{ma}$ as:

\begin{equation}
\hat{F}_{ma} = \hat{F}_c \odot A_{ma}.
\end{equation}

\begin{figure}[!t]
    \centerline{\includegraphics[width=0.5\columnwidth]{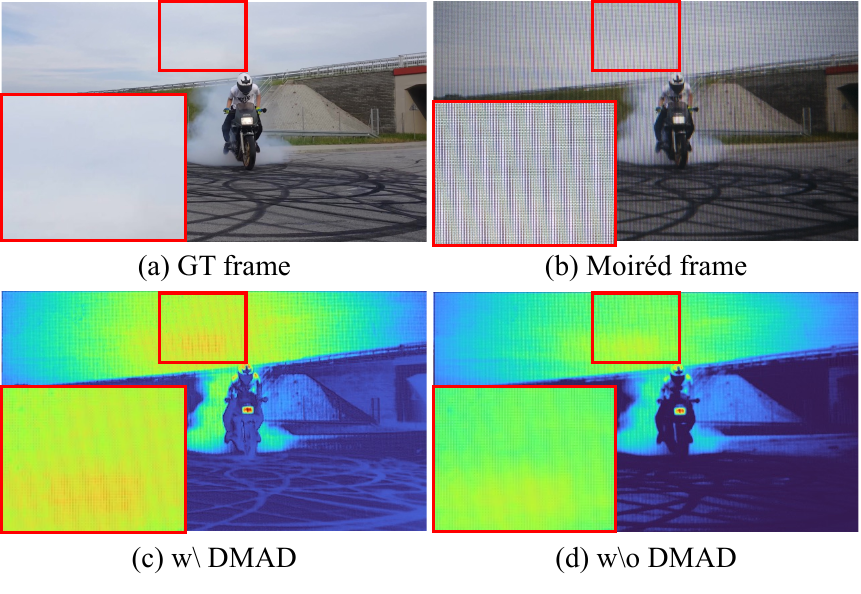}}
    \caption{Qualitative results of the proposed DMAD (Sec.~\ref{subsec_pm:dmad}) with the inclusion of MCB (Sec.~\ref{subsubsec_pm:MCB}). (a) shows the ground-truth frame, (b) the moiréd frame, (c) the feature map from the model with DMAD, and (d) the feature map from the model without DMAD. As observed in the highlighted red regions, (c) exhibits a stronger focus on the moiré patterns compared to (d).}
    \label{fig:dmad_attn} 
\end{figure}

Fig.~\ref{fig:dmad_attn}-(c) and (d) present the qualitative results of the moiré-adaptive feature map $\hat{F}_{ma}$. As shown in Fig.~\ref{fig:dmad_attn}, the model equipped with the proposed DMAD and MCB (Fig.~\ref{fig:dmad_attn}-(c)) exhibits a stronger focus on moiré patterns compared to the model without DMAD and MCB. This demonstrates that our approach effectively generates moiré-adaptive features.
The generated moiré-adaptive feature map $\hat{F}_{ma}$ allows the model to adaptively focus on regions with varying moiré intensity, helping to prevent over-smoothing of the input frames and supporting effective removal of moiré patterns while preserving fine details. 

\subsection{Spatio-Temporal Adaptive Demoiréing module (STAD)} \label{subsec_pm:stad}

\begin{figure}[!t]
    \centerline{\includegraphics[width=0.7\columnwidth]{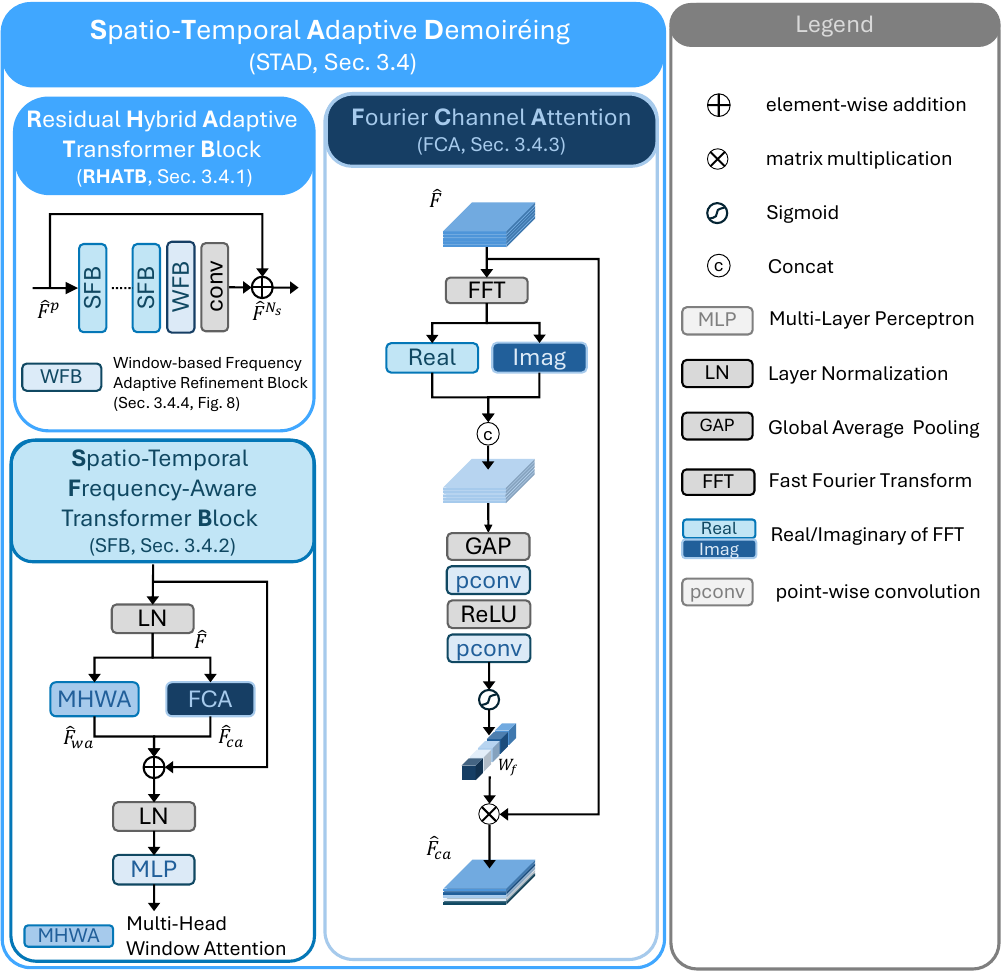}}
    \caption{Figure of the proposed Spatio-Temporal Adaptive Demoiréing (STAD, Sec.~\ref{subsec_pm:stad}). The STAD consists of $N_r$ Residual Hybrid Adaptive Transformer Blocks (RHATB, Sec.~\ref{subsubsec_pm:rhatb}), each of which is composed of $N_s$ Spatio-Temporal Frequency-Aware Transformer Blocks (SFB, Sec.~\ref{subsubsec_pm:sfb}). The SFB integrates Multi-Head Window Attention to simultaneously capture spatio-temporal information and Fourier Channel Attention (FCA, Sec.~\ref{subsubsec_pm:fca}) to adaptively process channel information in the frequency domain.}
    \label{fig:stad} 
\end{figure}

To incorporate global characteristics within a moiré‑contaminated frame, implicitly align adjacent frames, and enable information exchange across multiple moiré frames, we propose the \textbf{S}patio‑\textbf{T}emporal \textbf{A}daptive \textbf{D}emoiréing (STAD) module, which operates in conjunction with the moiré‑adaptive feature extracted in Sec.~\ref{subsec_pm:stad}. The STAD consists of $N_{r}$ \textbf{R}esidual \textbf{H}ybrid \textbf{A}daptive \textbf{T}ransformer \textbf{B}locks (RHATB) followed by a single convolutional layer as:

\begin{equation}
    \hat{F}_d = \mathtt{STAD}(\hat{F}_{ma}),\quad
    \mathtt{STAD} = conv\left(\left\{\mathtt{RHATB}_i\right\}_{i=1}^{N_r}\right).
\end{equation}

$\hat{F}_{ma}$ denotes the moiré-adaptive feature extracted from the DMAD (Sec.~\ref{subsec_pm:dmad}) module. $conv$ refers to a $3\times3$ convolutional layer, \texttt{STAD} and \texttt{RHATB} indicate the Spatio-Temporal Adaptive Demoiréing (STAD) module and the Residual Hybrid Adaptive Transformer Block (RHATB), respectively. $N_r$ represents the number of RHATB blocks used within the STAD module.

This architecture enables the model to simultaneously capture the global structure, local patterns, and temporal dynamics of moiré patterns in a unified manner.

\subsubsection{Residual Hybrid Adaptive Transformer Block (RHATB)} \label{subsubsec_pm:rhatb}
The proposed \textbf{R}esidual \textbf{H}ybrid \textbf{A}daptive \textbf{T}ransformer \textbf{B}lock (RHATB) takes the moiré-adaptive feature $\hat{F}_{ma}$ as input and is designed to jointly model spatial and temporal information. It is built upon an attention layer capable of capturing global context. RHATB consists of $N_s$ \textbf{S}patio-Temporal \textbf{F}requency-Aware \textbf{T}ransformer Blocks (SFB) as: 

\begin{equation}
\mathtt{RHATB}_i = \left\{\mathtt{SFB}_{j}\right\}_{j=1}^{N_s} ,
\end{equation}
followed by a \textbf{W}indow-based \textbf{F}requency Adaptive Refinement \textbf{B}lock (WFB), as shown in Fig.~\ref{fig:wfb}. $N_s$ denotes the number of SFB block. As discussed in Sec.~\ref{subsubsec_pm:frequency}, when applying FFT, the characteristics of moiré patterns can be separated in the frequency domain. Specifically, the structural properties of the moiré pattern are captured in the amplitude component, while surface-level artifacts reside in the phase component. Leveraging this property, we propose WFB, which explicitly separates and processes amplitude and phase information.
Detailed descriptions of SFB and WFB are provided in Sec.~\ref{subsubsec_pm:sfb} and Sec.~\ref{subsubsec_pm:wfb}, respectively.

\subsubsection{Spatio-Temporal Frequency-aware Transformer Block (SFB)} \label{subsubsec_pm:sfb}
As discussed in Sec.~\ref{subsubsec_pm:temporal}, moiré patterns can vary significantly even between adjacent frames. Additionally, as noted in Sec.~\ref{subsubsec_pm:moire_single_image}, their appearance also differs greatly within a single frame. Furthermore, as described in Sec.~\ref{subsubsec_pm:frequency}, the Fourier domain facilitates the separation and handling of moiré-specific characteristics. To leverage these properties, we propose the \textbf{S}patio-Temporal \textbf{F}requency-aware Transformer \textbf{B}lock (SFB).

The SFB first applies layer normalization, followed by a window attention mechanism, similar to the standard Swin Transformer \cite{swin}. Unlike typical window attention that partitions the spatial dimensions $(H, W)$ into $(k, k)$ windows, we adopt a 3D partitioning scheme, similar to video Swin Transformer variants \cite{vswin, psrt}, dividing the temporal and spatial dimensions into $(t, k, k)$ windows. This enables the \textbf{M}ulti-\textbf{H}ead \textbf{W}indow \textbf{A}ttention (MHWA) to simultaneously model both spatial and temporal dependencies.

In parallel, to adaptively modulate channel-wise representations in the frequency domain, we incorporate the proposed \textbf{F}ourier \textbf{C}hannel \textbf{A}ttention (FCA) module. Details of FCA are provided in Sec.~\ref{subsubsec_pm:fca}.

After jointly capturing spatio-temporal and frequency-aware channel features, the output is further refined through a layer normalization layer and a multi-layer perceptron (MLP), as commonly done in transformer architectures, to enhance the representational capacity of the features.

\subsubsection{Fourier Channel Attention (FCA)} \label{subsubsec_pm:fca}

In addition to the window attention operation introduced in Sec.~\ref{subsubsec_pm:sfb}, channel-wise interactions play an important role, particularly given that moiré patterns are more separable in the frequency domain. To leverage this property, we propose the \textbf{F}ourier \textbf{C}hannel \textbf{A}ttention (FCA) module. The FCA begins by applying FFT to the input feature $\hat{F}$.

\begin{align}
\mathcal{F}\{\hat{F}\}_{u,v} &= F^{\mathrm{re}}_{u,v} \;+\; \mathbf{i}\,F^{\mathrm{im}}_{u,v} \label{eq:fft},\\
F^{\mathrm{re}}_{u,v} &= \operatorname{Re}\bigl(\mathcal{F}\{\hat{F}\}_{u,v}\bigr) \label{eq:fft_real},\\
F^{\mathrm{im}}_{u,v} &= \operatorname{Im}\bigl(\mathcal{F}\{\hat{F}\}_{u,v}\bigr) \label{eq:fft_imag}.
\end{align}

Eq.~\ref{eq:fft} expresses the 2D Fourier transform of the spatial image $\hat{F}$ at frequency coordinate $(u,v)$ as a complex number with $\mathbf{i}=\sqrt{-1}$ denoting the imaginary unit. This complex number consists of a real part $F^{\mathrm{re}}_{u,v}$ and an imaginary part $F^{\mathrm{im}}_{u,v}$. Eq.~\ref{eq:fft_real} defines the real part $F^{\mathrm{re}}_{u,v}$ as the real component of the complex Fourier coefficient $\mathcal{F}\{\hat{F}\}_{u,v}$.
Eq.~\ref{eq:fft_imag} defines the imaginary part $F^{\mathrm{im}}_{u,v}$ as the imaginary component of the complex Fourier coefficient $\mathcal{F}\{\hat{F}\}_{u,v}$. The separated real-valued feature $F^{\mathrm{re}}_{u,v}$ and imaginary-valued feature $F^{\mathrm{im}}_{u,v}$ are concatenated along the channel dimension. This concatenated feature is then passed through a series of operations: global average pooling (GAP), point-wise convolution, ReLU activation, and another point-wise convolution, to produce the channel attention weight $W_f \in \mathbb{R}^{1 \times 1 \times C}$. The resulting attention weight is multiplied with the original input feature $\hat{F}$ to yield the channel-adaptively modulated feature $\hat{F}_{ca}$.
The resulting frequency channel-attended feature $\hat{F}_{ca}$ is scaled by a learnable weight $\lambda_{ca}$ and added to the output of the MHWA module, denoted as $\hat{F}_{wa}$:

\begin{equation}
\hat{F}_{a} = \hat{F}_{wa} + \lambda_{ca} \hat{F}_{ca}.
\end{equation}

This formulation allows the model to integrate both spatio-temporal information, captured by MHWA, and channel-wise frequency information, modulated by FCA, thereby enabling adaptive representation in the frequency domain.

\subsubsection{Window-based Frequency Adaptive Refinement Block (WFB)} \label{subsubsec_pm:wfb}

\begin{figure}[!t]
    \centerline{\includegraphics[width=0.6\columnwidth]{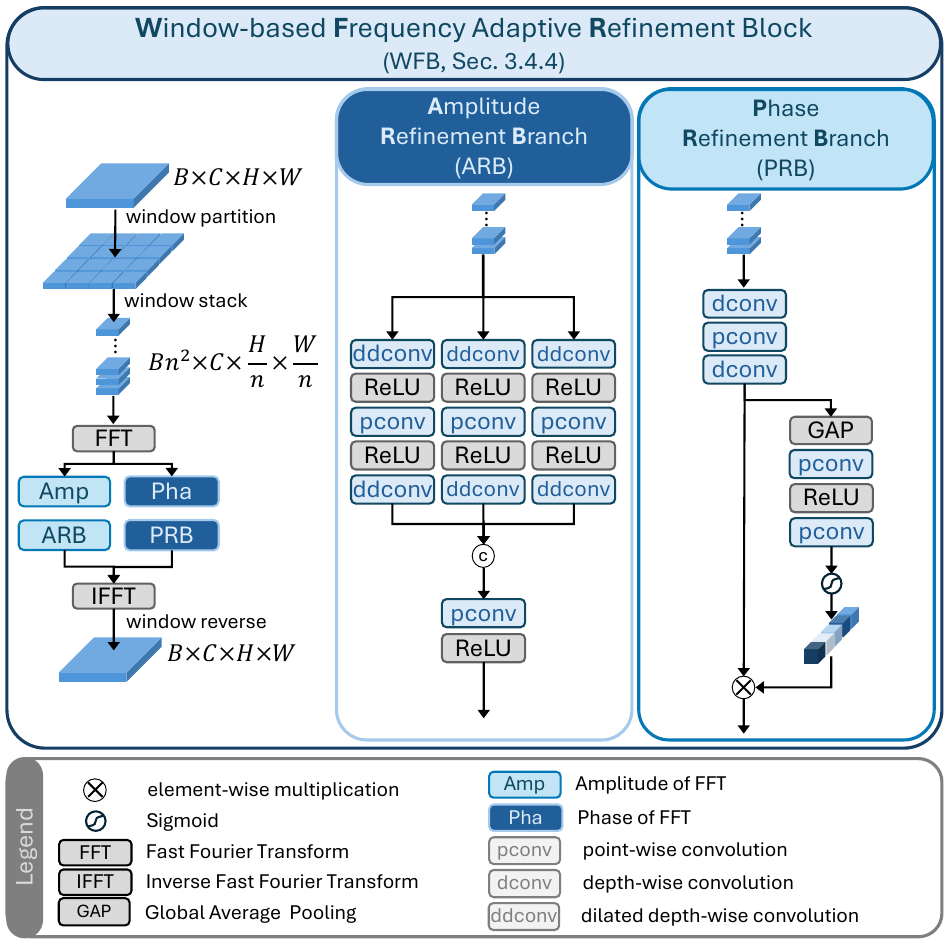}}
    \caption{Figure of the proposed Window-based Frequency Adaptive Refinement Block (WFB, Sec.~\ref{subsubsec_pm:wfb}). The WFB first partitions the input into windows, applies FFT to each window, and then processes the resulting amplitude and phase through the Amplitude Refinement Branch (ARB) and Phase Refinement Branch (PRB), respectively. This design enables the separation and handling of moiré pattern characteristics embedded in the amplitude and phase.}
    \label{fig:wfb} 
\end{figure}

As mentioned in Sec.~\ref{subsubsec_pm:frequency}, the amplitude and phase components obtained by applying the Fourier transform to a moiré-contaminated frame capture different characteristics of the moiré pattern. To refine the features extracted by the SFB block (Sec.~\ref{subsubsec_pm:sfb}) in the frequency domain, we propose the \textbf{W}indow-based \textbf{F}requency Adaptive Refinement \textbf{B}lock (WFB), as shown in Fig.~\ref{fig:wfb}.
To maintain consistency with the window attention operations in the SFB block, we apply window partitioning to the features before processing. This localized operation ensures that each window is processed independently, which improves training stability, as suggested in \cite{uformer}. We then apply FFT to each windowed feature and decompose it into amplitude and phase components as:

\begin{align}
A_{u,v} &= \sqrt{\bigl(F^{\mathrm{re}}_{u,v}\bigr)^{2} + \bigl(F^{\mathrm{im}}_{u,v}\bigr)^{2}} \label{eq:amp},\\
\Phi_{u,v} &= {\tan}^{-1}\!\bigl(F^{\mathrm{im}}_{u,v}/F^{\mathrm{re}}_{u,v}\bigr) \label{eq:phase}.
\end{align}

In Eq.~\eqref{eq:amp}-\eqref{eq:phase}, $\hat{F}$ denotes the spatial‑domain feature map under analysis, while $\mathcal{F}\{\hat{F}\}_{u,v}$ represents its 2‑D discrete Fourier transform evaluated at the horizontal and vertical frequency indices $(u,v)$.  This complex coefficient is decomposed into a real part $F^{\mathrm{re}}_{u,v}$ and an imaginary part $F^{\mathrm{im}}_{u,v}$, with $\mathbf{i}=\sqrt{-1}$ denoting the imaginary unit.  The amplitude \(A_{u,v}\) in \eqref{eq:amp} is the Euclidean magnitude of this complex number, capturing the energy of the corresponding frequency component, whereas the phase \(\Phi_{u,v}\) in \eqref{eq:phase} is its argument, obtained via the two‑argument arctangent $\operatorname{\tan^{-1}}(\cdot)$ and encoding the relative positional information of the signal content across the spatial domain.

For each branch of the transformed amplitude and phase, we apply the Amplitude Refinement Block (ARB) and Phase Refinement Block (PRB). As discussed in Sec.~\ref{subsubsec_pm:frequency}, the ARB employs three dilated convolutional layers with different receptive fields to capture the structural characteristics of moiré patterns, which are widely distributed in the amplitude. Specifically, one branch utilizes a standard $3 \times 3$ depth-wise convolution, while the other two branches adopt dilated depth-wise convolutions with dilation rates of 2 and 4, respectively. Since channel interactions in the amplitude are also crucial, a $1 \times 1$ point-wise convolution is inserted between the depth-wise convolutions. The outputs of the three branches are then concatenated and fused through another $1 \times 1$ convolution layer to integrate the information from the three receptive fields.

In the PRB, as described in Sec.~\ref{subsubsec_pm:frequency}, we leverage the fact that the phase encodes appearance information, including the color characteristics and smoothness of moiré patterns. To exploit this property, we apply depthwise-separable convolutions \cite{dsconv} followed by an SEBlock \cite{senet} to adaptively utilize channel-wise information.

The refined amplitude and phase features are subsequently transformed back into the spatial domain via an IFFT and passed to the next module. Through the WFB, the model can separately process the distinct characteristics of moiré patterns contained in the amplitude and phase, thereby enabling effective utilization of their frequency-domain representations.

\subsection{Training Strategy} \label{subsec_pm:training}

To optimize the model effectively, we adopt a multi-stage training strategy. In stage 1, as shown in Fig.~\ref{fig:framework_overview}, 
only the STAD module composed of the SFE and EIB components are trained. During this stage, the model is optimized using a combination of L1 loss and perceptual loss:

\begin{equation}
\mathcal{L}_{\mathrm{stg1}} = \, \mathcal{L}_{\mathrm{L1}} + \lambda_1 \, \mathcal{L}_{\mathrm{percept}}.
\label{eq:total_loss}
\end{equation}

The perceptual loss $\mathcal{L}_{\mathrm{percept}}$ evaluates the perceptual quality by utilizing the feature map obtained from the $j$-th convolution following the activation before the $i$-th max-pooling layer within a pre-trained VGG-19 backbone. The perceptual loss is defined as:

$$
\mathcal{L}_{\mathrm{percept}}
= \frac{1}{C \, H \, W}
\Bigl\|
\phi_{i,j}\bigl(\hat{I}\bigr) - \phi_{i,j}\bigl(I\bigr)
\Bigr\|_2^{2},
$$
where $\phi_{l}$ denotes the feature extracted from the \texttt{relu5\_4} layer of VGG-19 \cite{vggnet}, $\hat{I}$ represents the restored image, and $I$ is the reference (ground-truth) image. $C_l$, $H_l$, and $W_l$ indicate the number of channels and the spatial resolution of the corresponding feature map, respectively.

This initial phase enables the STAD module to capture spatio-temporal representations, which are critical for modeling moiré patterns in the temporal domain.

In stage 2, we integrate the DMAD module to generate moiré-adaptive features and perform joint training with the STAD module. Since the STAD has already learned the spatio-temporal characteristics of moiré patterns during Stage 1, its learning rate is scaled down to $1/10$ of the original value to stabilize joint optimization. To optimize DMAD, we additionally incorporate three task-specific losses of DDB, as follows:

\begin{equation}
\mathcal{L}_\mathrm{DDA}=
\lambda_2 \mathcal{L}_{\mathrm{PD}} 
+ \lambda_3 \mathcal{L}_{\mathrm{MP}} 
+ \lambda_4 \mathcal{L}_{\mathrm{MC}} \label{eq:total_loss},
\end{equation}

\begin{equation}
\mathcal{L}_{\mathrm{stg2}} = \mathcal{L}_{\mathrm{stg1}} + \mathcal{L}_{\mathrm{DDA}}.
\end{equation}

This stage encourages the model to focus on generating features that are adaptive to moiré patterns, guided by the structural, predictive, and decoupling constraints introduced by the DMAD module.

%% file: sections/Experiments.tex
\section{Experiments}

\subsection{Experiment Setting}
\subsubsection{Implementation Details} \label{subsubsec_exp:implement}
Our video demoiréing model takes three consecutive RAW frames as input. The number of RHATB blocks $N_r$ is set to 4, while the number of SFB blocks $N_s$ is set to 5. For the STAD module, the window size $(t,k,k)$ is configured as $(2,7,7)$. During training, frames are cropped to $128 \times 128$ for RAW inputs and $256 \times 256$ for sRGB inputs. In stage 1, we employ the AdamW optimizer with $\beta_1 = 0.9$ and $\beta_2 = 0.999$, and the initial learning rate is set to $1 \times 10^{-4}$ to training STAD (Sec.~\ref{subsec_pm:stad}), SFE and EIB. The hyper-parameter $\lambda_1$ for the loss function in stage 1 is set to 0.01. In stage 2, we also use AdamW \cite{adamw} to optimize the STAD, SFE, and EIB modules, initializing the learning rate to $1 \times 10^{-5}$ since the model is loaded from stage 1. A separate optimizer with AdamW ($\beta_1 = 0.9$, $\beta_2 = 0.999$) is employed for effective training of the DMAD (Sec.~\ref{subsec_pm:dmad}). The loss functions in stage 2 include the L1 and perceptual losses from stage 1, along with PD loss ($\mathcal{L}_\mathrm{MC}$), MC loss ($\mathcal{L}_\mathrm{MC}$), and MP loss ($\mathcal{L}_\mathrm{MP}$) for optimizing DMAD. The corresponding weights $\lambda_2, \lambda_3, \lambda_4$ are set to 1, 1, and 0.5, respectively. For both stage 1 and stage 2, the batch size is set to 3 and training is conducted for a total of 50 epochs. All models are trained on a single NVIDIA RTX 6000 Ada GPU.

\subsubsection{Datasets} \label{subsubsec_exp:implement}
To evaluate the performance of the proposed MoCHA-former, we utilize the RawVDemoiré dataset \cite{rawvdemoire}. This dataset consists of clean sRGB video clips, moiréd sRGB video clips, moiréd RAW video clips, and (pseudo) clean RAW video clips converted from sRGB video clips. The RawVDemoiré dataset contains a total of 300 video clips, divided into 250 training clips and 50 test clips. Each video clip comprises 60 frames; the sRGB frames have a resolution of $1280 \times 720$, while the RAW frames are provided in a packed format (RGGB bayer filter) with a resolution of $640 \times 360$ and 4 channels.

To further validate the generalization capability of our model, we also employ the VDemoiré dataset \cite{vdemoire}, which consists solely of sRGB video clips. This dataset contains paired clean and moiréd sRGB video clips, comprising a total of 290 pairs. Each video clip includes 60 frames with a resolution of $1280 \times 720$. To evaluate our model, we train our model from scratch using two datasets.

\subsection{Comparison to State-of-the-Art Methods}

To demonstrate the outperforming performance of the proposed MoCHA-former, we compare our approach with previous state-of-the-art (SOTA) image and video demoiréing models on the RawVDemoiré dataset \cite{rawvdemoire} and the VDemoiré dataset \cite{vdemoire}. Table~\ref{tab:rawvdemoire} reports the comparison against image demoiréing models RDNet \cite{tmm22} and RRID \cite{rrid}, as well as video demoiréing models VDemoire \cite{vdemoire}, DTNet \cite{dtnet}, and RawVDemoire \cite{rawvdemoire}.
The superscript “*” denotes that the corresponding model was originally trained on sRGB images and has been re-trained from scratch on RAW data to ensure a fair comparison in the raw domain. Our approach outperforms the second-best model (RawVDemoire) by \mbox{$+\Delta{\text{1.001}}$\,dB} in PSNR$\uparrow$, \mbox{$+\Delta{\text{0.0078}}$} in SSIM$\uparrow$, and \mbox{$-\Delta{\text{0.0079}}$} in LPIPS$\downarrow$. Moreover, our network contains only \textbf{3.64\,M} parameters, making it considerably more compact than RawVDemire (6.59\,M, +80\% than Ours) and competitive with lighter image-domain models, yet still delivering the best overall accuracy. 

To verify that the proposed model also achieves competitive performance in the sRGB domain, we provide a comparison on the VDemoiré dataset. Table~\ref{tab:vdemoire} reports results against sRGB image demoiréing models MBCNN \cite{8} and WDNet \cite{6}, as well as sRGB video demoiréing models VDemoiré and DTNet \cite{dtnet}. 
Compared to the second-best model (DTNet), our approach improves PSNR$\uparrow$ by $+0.009$ and SSIM$\uparrow$ by $+0.005$, while reducing LPIPS$\downarrow$ by $0.002$. Moreover, it uses only $48.9\%$ of the parameters, demonstrating higher parameter efficiency.

\begin{figure}[!t]
  \centering
  \centerline{\includegraphics[width=\columnwidth]{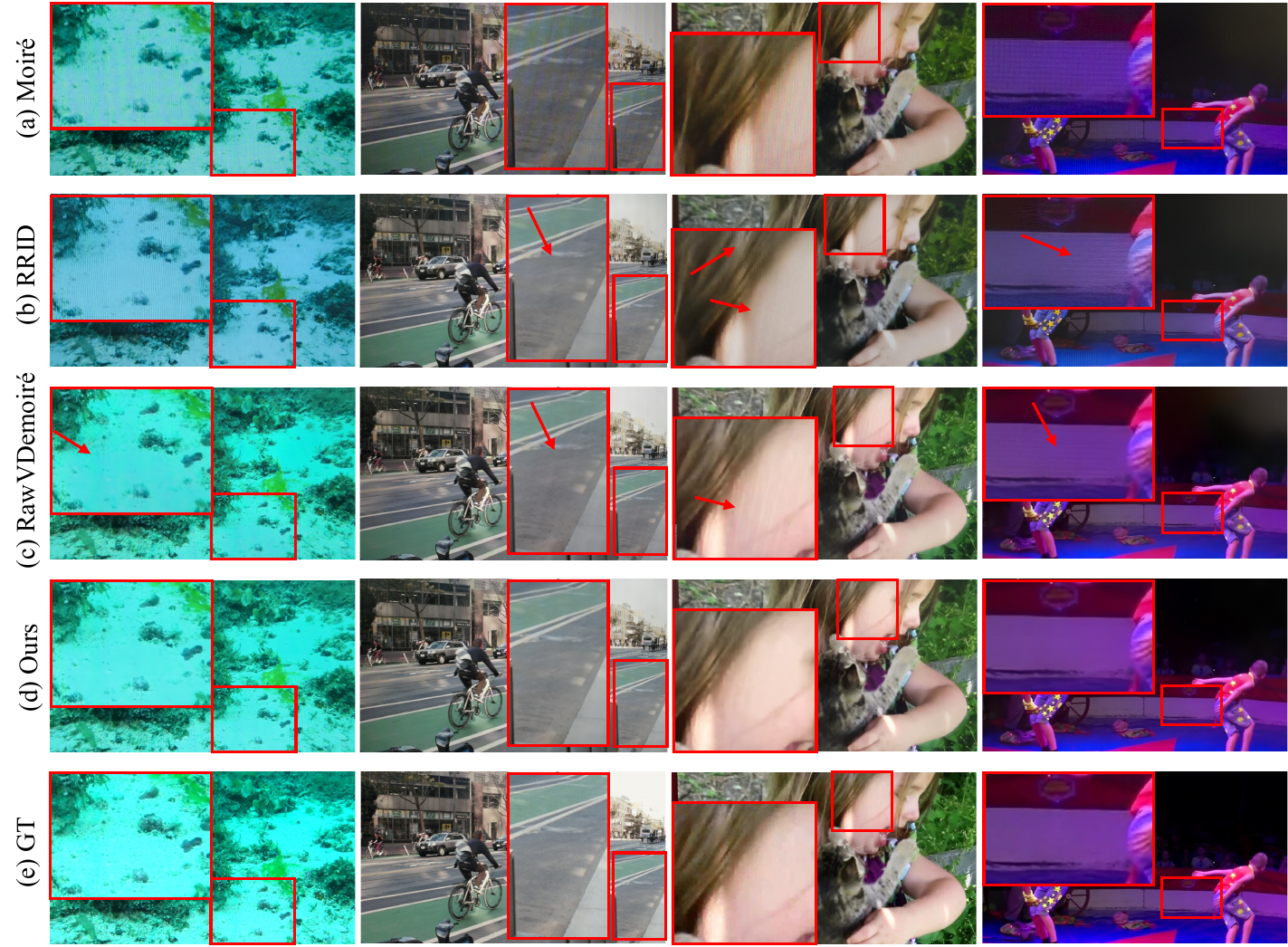}}
  \caption{Qualitative results on the RawVDemoiré dataset~\cite{rawvdemoire}. (a) shows the input moiréd frame, (e) is the corresponding ground-truth frame, and (b)–(d) present the results of RRID~\cite{rrid}, RawVDemoiré~\cite{rawvdemoire}, and MoCHA-former (Ours), respectively. As highlighted by the red boxes and arrows in (b), RRID fails to fully resolve the color distortion compared to Ours and retains stripe-like moiré patterns. Similarly, as indicated by the red boxes and arrows in (c), RawVDemoiré does not completely remove grid- or stripe-like moiré artifacts. In contrast, Ours effectively removes both grid and stripe moiré patterns while also restoring accurate color appearance.}
  \label{fig:main_qual_raw}
\end{figure}

\begin{figure}
  \centering
  \centerline{\includegraphics[width=\columnwidth]{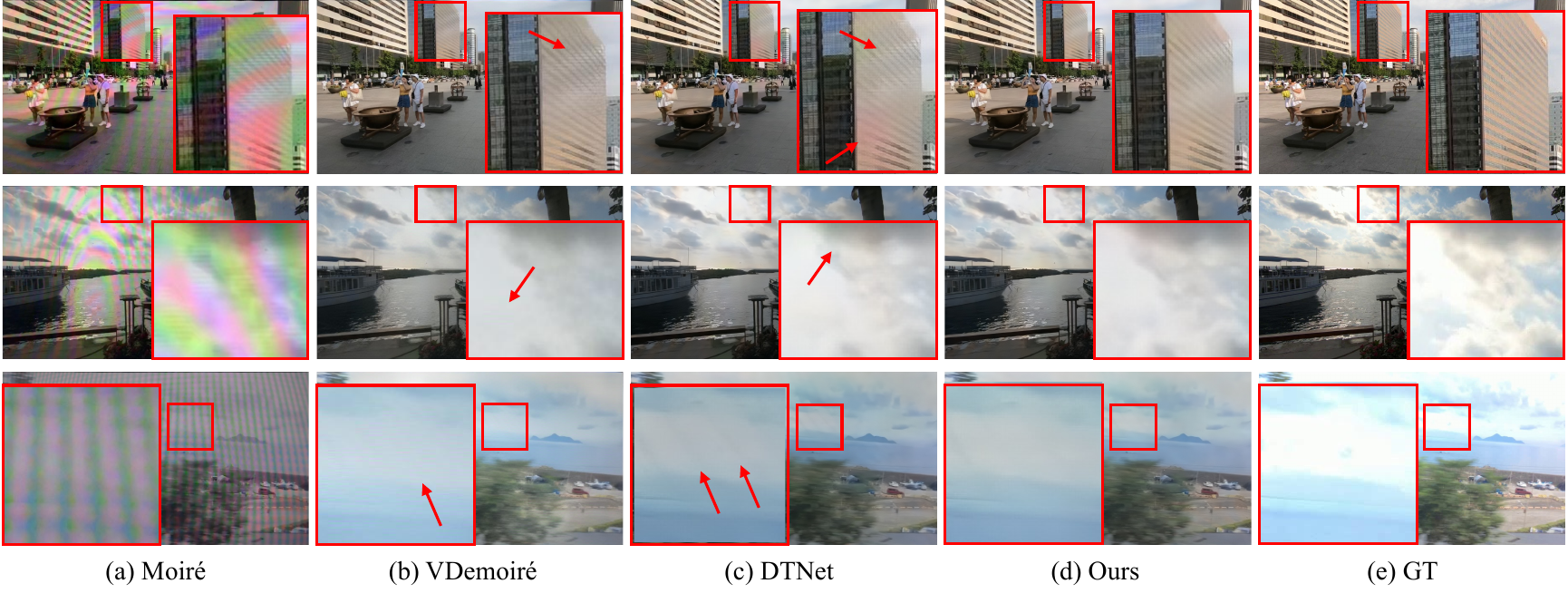}}
    \caption{Qualitative results on the VDemoiré dataset~\cite{vdemoire}. (a) shows the input moiréd frame, (e) is the corresponding ground-truth frame, and (b)–(d) present the results of VDemoiré~\cite{vdemoire}, DTNet~\cite{dtnet}, and MoCHA-former (Ours), respectively. As observed in (b), VDemoiré fails to resolve the color distortion across the entire frame compared to Ours. In (c), DTNet alleviates the overall color distortion; however, as highlighted by the red boxes and arrows, localized color artifacts from moiré patterns remain, and stripe-shaped moiré structures are still present. In contrast, Ours removes most of the stripe-shaped moiré patterns while simultaneously restoring the correct color appearance.}
    \label{fig:main_qual_rgb} 
\end{figure}

We present qualitative results on the RawVDemoiré and VDemoiré datasets in Fig.~\ref{fig:main_qual_raw} and Fig.~\ref{fig:main_qual_rgb}, respectively. 
In Fig.~\ref{fig:main_qual_raw}, (a) shows the input moiréd frame, (e) is the corresponding ground-truth frame, and (b)–(d) are the output frames from RRID~\cite{rrid}, RawVDemoiré~\cite{rawvdemoire}, and MoCHA-former (Ours), respectively. The red boxes in Fig.~\ref{fig:main_qual_raw}-(b) show that RRID fails to completely remove the grid-like structures of moiré patterns across all rows. Fig.~\ref{fig:main_qual_raw}-(c) removes more of the grid-like artifacts compared to Fig.~\ref{fig:main_qual_raw}-(b), but still retains localized color distortions.

In Fig.~\ref{fig:main_qual_rgb}, (a) shows the input moiréd frame, (e) is the corresponding ground-truth frame, and (b)–(d) are the output frames from VDemoiré~\cite{vdemoire}, DTNet~\cite{dtnet}, and MoCHA-former (Ours), respectively. The red boxes in Fig.~\ref{fig:main_qual_rgb}-(b) reveal that VDemoiré fails to resolve the overall color distortion compared to Ours and still retains grid-like moiré structures. In Fig.~\ref{fig:main_qual_rgb}-(c), DTNet preserves the overall color tone better than VDemoiré but still exhibits localized color distortions characteristic of moiré patterns.

\begin{table}[!t]
    \centering
    \begin{tabular}{c l c c c c c}
        \hline
         & Method & PSNR$\uparrow$ & SSIM$\uparrow$ & LPIPS$\downarrow$ & Params (M)\\
        \hline
        \multirow{2}{*}{\textbf{Image}} 
        & RDNet (TMM'22)~\cite{tmm22}  & 25.892 & 0.8939 & 0.1508 & 2.72 \\
        & RRID (ECCV'24)~\cite{rrid}  & 27.283 & 0.9029 & 0.1168 & 2.37 \\
        \hline
        \multirow{6}{*}{\textbf{Video}} 
        & VDMoiré (CVPR'22)~\cite{vdemoire} & 27.277 & 0.9071 & 0.1044 & 5.84 \\
        & VDMoiré* (CVPR'22)~\cite{vdemoire} & 27.747 & 0.9116 & 0.0995 & 5.84 \\
        & DTNet (AAAI'24)~\cite{dtnet} & 27.363 & 0.8963 & 0.1425 & 3.99 \\
        & RawVDmoiré (NeurIPS'24)~\cite{rawvdemoire} & \textcolor{blue}{\underline{28.706}} & \textcolor{blue}{\underline{0.9201}} & \textcolor{blue}{\underline{0.0904}} & 6.59 \\
        & MoCHA-former (\textbf{Ours})& \textcolor{red}{\textbf{29.707}} & \textcolor{red}{\textbf{0.9279}} & \textcolor{red}{\textbf{0.0825}} & 3.64 \\
        \hline
    \end{tabular}
    \caption{Comparison of image and video models in RAW video demoiréing dataset \cite{rawvdemoire}. The symbol ``$*$" denotes a model originally designed for processing sRGB images, which has been modified and retrained to handle RAW inputs.}
    \label{tab:rawvdemoire}
\end{table}
\begin{table}[!t]
    \centering
    \begin{tabular}{c l c c c c c}
        \hline
         & Method & PSNR$\uparrow$ & SSIM$\uparrow$ & LPIPS$\downarrow$ & Params (M)\\
        \hline
        \multirow{2}{*}{\textbf{Image}} 
        & MBCNN (CVPR'20)~\cite{8} & 24.060 & 0.849 & 0.211 & 14.19 \\
        & WDNet (ECCV'20)~\cite{6} & 23.971 & 0.834 & 0.205 & 17.26 \\
        \hline
        \multirow{3}{*}{\textbf{Video}} 
        & VDemoiré (CVPR'22)~\cite{vdemoire} &                             25.230 &  \textcolor{red}{\textbf{0.860}}&   0.157 & 5.84 \\
        & DTNet (AAAI'24)~\cite{dtnet}       & \textcolor{blue}{\underline{26.503}}&    0.854 & \textcolor{red}{\textbf{0.149}}& 7.36 \\
        & MoCHA-former (\textbf{Ours}) &  \textcolor{red}{\textbf{26.512}}& \textcolor{blue}{\underline{0.859}}&  \textcolor{blue}{\underline{0.151}}& 3.60 \\
        \hline
    \end{tabular}
    \caption{Comparison of image and video models in sRGB video demoiréing dataset \cite{vdemoire}}
    \label{tab:vdemoire}
\end{table}

\begin{table}[t]
  \centering
  \begin{tabular}{@{}lccc@{}}           
    \toprule
    \textbf{Depths} & \textbf{PSNR$\uparrow$} & \textbf{SSIM$\uparrow$} & \textbf{LPIPS$\downarrow$} \\ \midrule
    $[2,2,2]$      &                             28.4847 &                              0.9178 &                              0.0971 \\
    $[3,3,3]$      &                             28.5211 &                              0.9185 &                              0.0952 \\
    $[4,4,4,4]$    & \textcolor{blue}{\underline{28.8692}}& \textcolor{blue}{\underline{0.9198}}& \textcolor{blue}{\underline{0.0892}} \\
    $[5,5,5,5]$    &     \textcolor{red}{\textbf{29.7071}}&     \textcolor{red}{\textbf{0.9279}}&     \textcolor{red}{\textbf{0.0825}} \\
    \bottomrule
  \end{tabular}
  \caption{Performance according to network depth. A longer array indicates a larger number of RHATB blocks ($N_r$), while higher values within the array correspond to an increased number of SFB modules ($N_s$). As observed above, the model performance improves proportionally as both $N_r$ and $N_s$ increase.}
  \label{tab:depth_performance}
\end{table}

\subsection{Ablation Study}

To validate the effectiveness of the proposed MoCHA-former, we conduct a series of ablation studies. 
First, to evaluate the contribution of each component, we compare the sub-modules of DMAD and STAD in Sec.~\ref{subsubsec_exp:model_arch_ablation}. 
Second, we demonstrate the superiority of STAD by comparing its implicit alignment with conventional explicit alignment methods. 
Third, to assess the ability of MDB-separated moiré patterns and content in generating moiré-adaptive features, we investigate different feature fusion strategies in Sec.~\ref{subsubsec_exp:mcb_ablation}. 
Fourth, to verify the effectiveness of the proposed DDB, we compare different loss functions in Sec.~\ref{subsubsec_exp:ddb_ablation}. 
Finally, we compare the performance of the proposed two-stage training strategy with that of a single-stage approach in Sec.~\ref{subsubsec_exp:training_strategy}.

All experiments are conducted on the RawVDemoiré dataset \cite{rawvdemoire}. For experimental efficiency, we reduce the number of RHATB and SFB blocks in our model during ablation. As observed in Table~\ref{tab:depth_performance}, the performance scales linearly with the number of RHATB and SFB blocks, which justifies this choice for efficient evaluation. The results of these ablation studies confirm that the proposed modules and configurations of MoCHA-former constitute an optimal design.

\subsubsection{Ablation Study on Model Architecture} \label{subsubsec_exp:model_arch_ablation}

\begin{figure}[!t]
    \centerline{\includegraphics[width=\columnwidth]{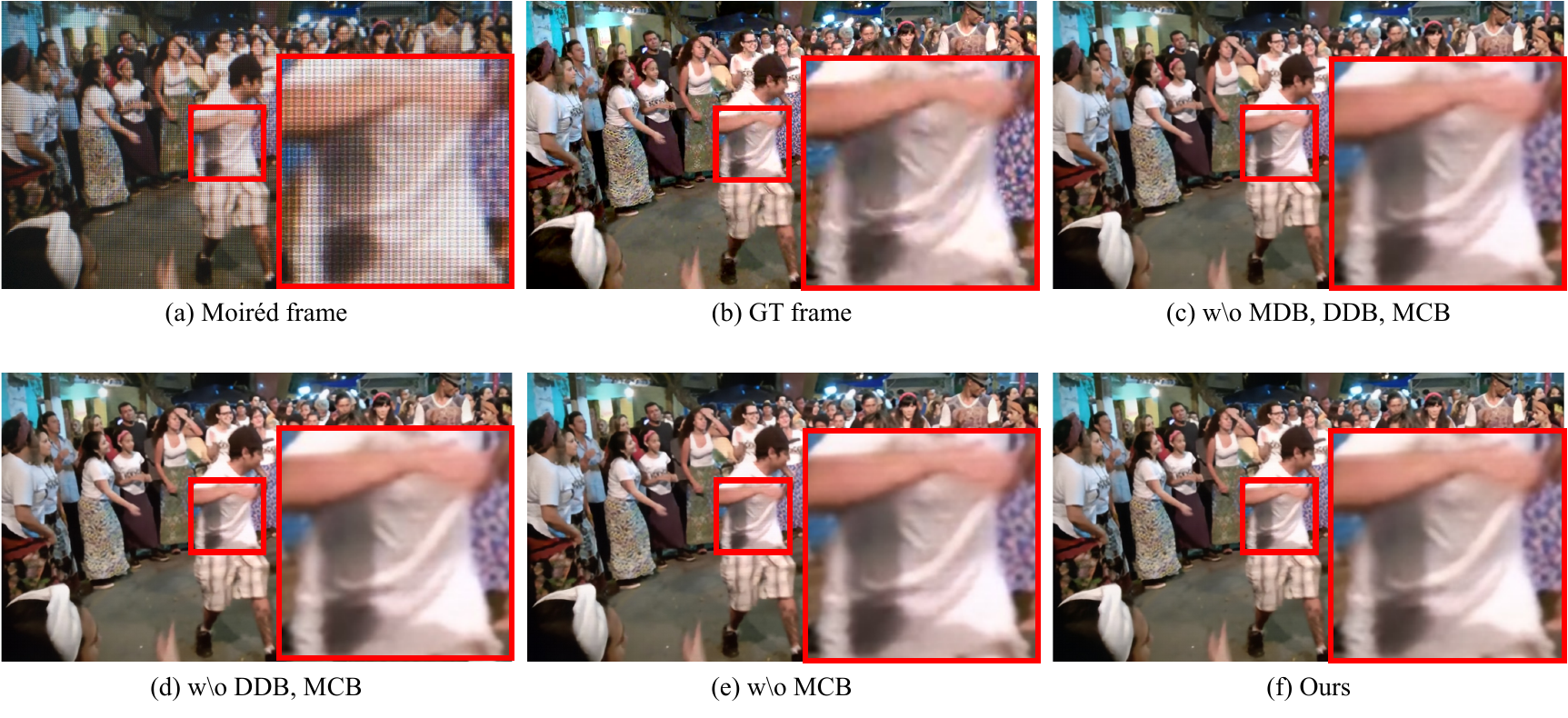}}
    \caption{Qualitative results for evaluating the DMAD module. (a) shows the input moiréd frame, (b) is the corresponding ground-truth frame, (c) is the model with all DMAD modules removed, (d) is the model with DDB and MCB removed from DMAD, (e) is the model with only MCB removed from DMAD, and (f) is the model with the full DMAD applied. All models are retrained from scratch. As shown in red box, Compared to (f), the results in (c)–(e) retain noticeable grid-like moiré structures.}
    \label{fig:main_arch_ablation} 
\end{figure}

The proposed MoCHA-former is composed of two key components: 1) DMAD and 2) STAD. The DMAD module is further divided into (i) MDB, (ii) DDB, and (iii) MCB, while the STAD module consists of (i) RHATB, (ii) FCA, and (iii) WFB. We conducted ablation studies with various combinations of these components to evaluate their effectiveness.

Table~\ref{tab:ab_architecture} presents the performance of the DMAD module. As shown in Table~\ref{tab:ab_architecture}, adding each proposed component progressively improves performance across all three metrics (PSNR$\uparrow$, SSIM$\uparrow$, LPIPS$\downarrow$). Fig.~\ref{fig:main_arch_ablation} provides qualitative results for each sub-module of DMAD. In Fig.~\ref{fig:main_arch_ablation}-(c)-(e), the characteristic grid-like moiré patterns remain more visible compared to our full model, demonstrating the importance of the proposed modules.

Furthermore, as shown in Table~\ref{tab:stad}, compared to using a vanilla Video Swin Transformer (Table~\ref{tab:stad}-(a)), sequentially incorporating FCA and WFB leads to consistent improvements in all three metrics (PSNR$\uparrow$, SSIM$\uparrow$, LPIPS$\downarrow$). For ablation results on the temporal alignment capability of STAD, please refer to Sec.~\ref{subsubsec_exp:alignment}.

\begin{table}[!t]
  \centering
  \begin{tabular}{c ccc ccc} 
    \toprule
    \multicolumn{1}{c}{} &
    \multicolumn{3}{c}{\textbf{Components}} &
    \multicolumn{3}{c}{\textbf{Metrics}} \\
    \cmidrule(lr){2-4}\cmidrule(l){5-7}
    & MDB & DDB & MCB & PSNR$\uparrow$ & SSIM$\uparrow$ & LPIPS$\downarrow$ \\ \midrule
    (a) & X & X & X &                  27.9969 &                  0.9149 &                  0.1007 \\
    (b) & O & X & X &                  28.3946 &                  0.9177 &                  0.0967 \\
    (c) & O & O & X & \textcolor{blue}{\underline{28.4401}}& \textcolor{blue}{\underline{0.9178}}& \textcolor{blue}{\underline{0.0965}}\\
    (d) & O & O & O &  \textcolor{red}{\textbf{28.5211}}&  \textcolor{red}{\textbf{0.9185}}&  \textcolor{red}{\textbf{0.0952}}\\
    \bottomrule
  \end{tabular}
  \caption{Ablation study of key components of DMAD. (a) removes the DMAD module, (b) adds only the MDB component, (c) further includes the DDB module, and (d) incorporates the full DMAD module. The best performance is achieved when all components of DMAD are employed.}
  \label{tab:ab_architecture}
\end{table}

\begin{table}[!t]
  \centering
  \begin{tabular}{c cc ccc} 
    \toprule
           & \multicolumn{2}{c}{\textbf{Components}}
           & \multicolumn{3}{c}{\textbf{Metrics}} \\
    \cmidrule(lr){2-3}\cmidrule(lr){4-6}
           & FCA & WFB
           & PSNR$\uparrow$ & SSIM$\uparrow$ & LPIPS$\downarrow$ \\ \midrule
    (a)    & X & X &                             28.4564  & \textcolor{blue}{\underline{0.9168}}&                            0.0955   \\
    (b)    & X & O & \textcolor{blue}{\underline{28.4719}}&                             0.9163  & \textcolor{blue}{\underline{0.0951}}\\
    (c)    & O & O &     \textcolor{red}{\textbf{28.5211}}&     \textcolor{red}{\textbf{0.9185}}&     \textcolor{red}{\textbf{0.0952}}\\
    \bottomrule
  \end{tabular}
  \caption{Ablation study on the components of STAD. (a) removes both the FCA and WFB modules, (b) adds only the WFB module, and (c) includes all components of STAD. All experiments were retrained starting from stage 1, and the reported results correspond to the final model after completing stage 2.}
  \label{tab:stad}
\end{table}

\subsubsection{Ablation Study on Decoupling Distillation Block (DDB)} \label{subsubsec_exp:ddb_ablation}

\begin{figure}[!t]
    \centerline{\includegraphics[width=\columnwidth]{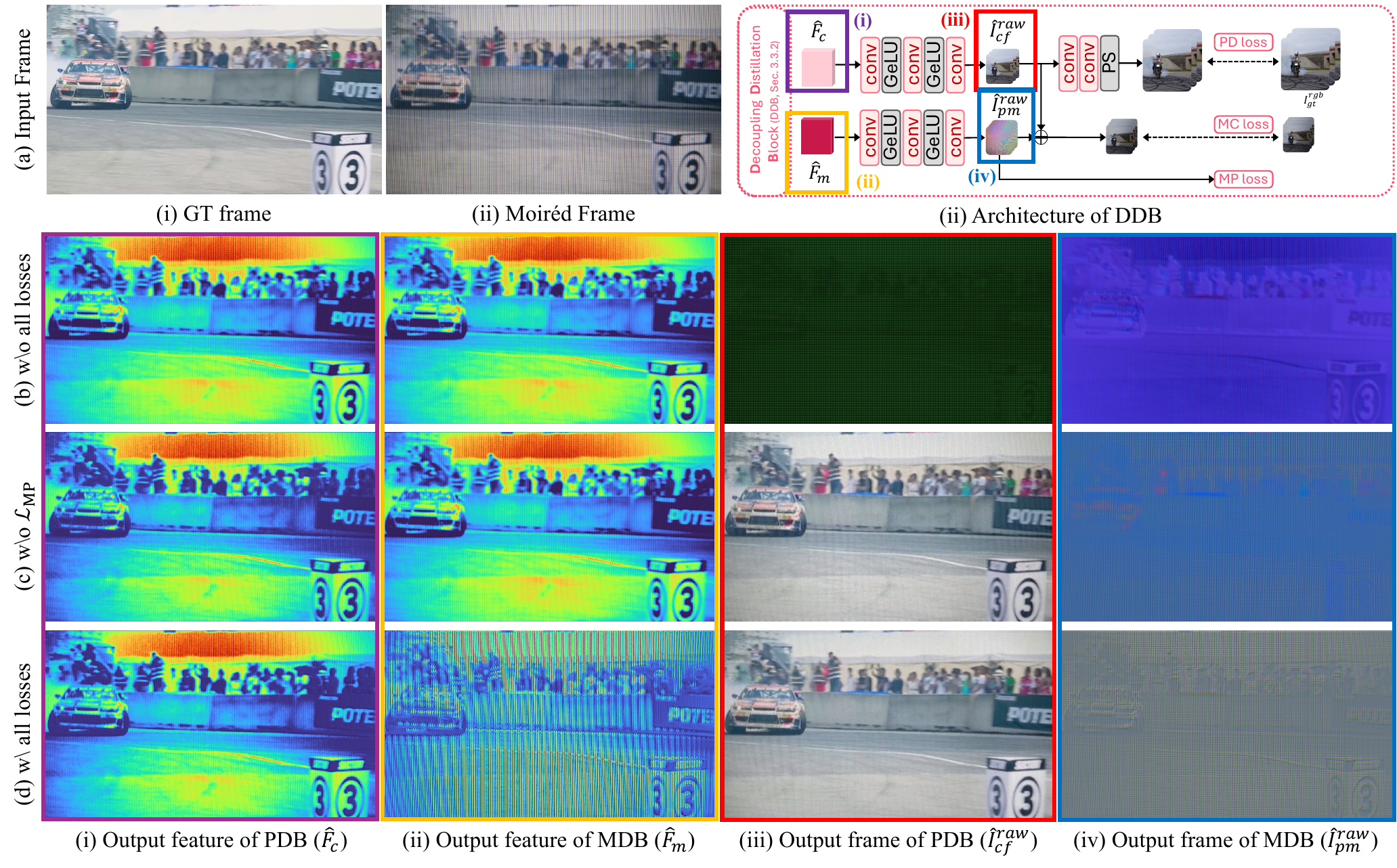}}
    \caption{Qualitative results illustrating the effects of each loss term applied to the DDB. Row (a) presents the ground-truth frame, the input frame, and the architecture of the DDB, respectively. In rows (b)–(d), column (iii) shows the pre-demoiréd frame, column (iv) depicts the predicted moiré pattern, column (i) presents the pre-demoiréd feature, and column (ii) shows the predicted moiré feature. In (b), moiré patterns and content are not properly disentangled at either the frame or feature level. In (c), disentanglement is achieved at the frame level but not at the feature level. In contrast, (d) demonstrates successful disentanglement at both the frame and feature levels.}
    \label{fig:ddb_ablation} 
\end{figure}

\begin{table}
  \centering
  \begin{tabular}{c ccc ccc} 
    \toprule
    \multicolumn{1}{c}{} &
    \multicolumn{3}{c}{\textbf{Components}} &
    \multicolumn{3}{c}{\textbf{Metrics}} \\
    \cmidrule(lr){2-4}\cmidrule(l){5-7}
        & $\mathcal{L}_\mathrm{PD}$& $\mathcal{L}_\mathrm{MC}$& $\mathcal{L}_\mathrm{MP}$& PSNR$\uparrow$ & SSIM$\uparrow$ & LPIPS$\downarrow$ \\ \midrule
    (a) & X & X & X &                             28.0024 &                              0.9160 &                              0.0998  \\
    (b) & O & O & X & \textcolor{blue}{\underline{28.4923}}& \textcolor{blue}{\underline{0.9169}}& \textcolor{blue}{\underline{0.0955}} \\
    (c) & O & O & O &     \textcolor{red}{\textbf{28.5211}}&     \textcolor{red}{\textbf{0.9185}}&     \textcolor{red}{\textbf{0.0952}} \\
    \bottomrule
  \end{tabular}
  \caption{Ablation study on the losses applied to the DDB. (a) retains the DDB structure but removes all associated losses, (b) excludes only the moiré prior (MP) loss from the DDB loss terms, and (c) employs all loss functions for DDB.}
  \label{tab:ddb_ablation}
\end{table}

Table~\ref{tab:ddb_ablation} reports the performance of each loss used in the DDB for separating content and moiré patterns, while Fig.~\ref{fig:ddb_ablation} provides visualizations of the separated content and moiré patterns in both the pixel and feature domains. Table~\ref{tab:ddb_ablation}-(a) corresponds to a model without any DDB-related losses. As shown, removing the DDB losses results in significantly lower PSNR$\uparrow$, SSIM$\uparrow$, and higher LPIPS$\downarrow$ values. Furthermore, as shown in Fig.~\ref{fig:ddb_ablation}-(b), the absence of DDB losses causes the model to output mixed features without properly separating the moiré patterns and content.
Table~\ref{tab:ddb_ablation}-(b) presents the model with PD and MC losses added. Compared to Table~\ref{tab:ddb_ablation}-(a), a substantial improvement in performance is observed. Consistently, Fig.~\ref{fig:ddb_ablation}-(c) shows that the moiré patterns and content are indeed separated; however, as seen in Fig.~\ref{fig:ddb_ablation}-(c)-(i) and (ii), the separation in the feature domain remains incomplete.
Finally, Table~\ref{tab:ddb_ablation}-(c) shows the model with all losses applied to the DDB. This configuration achieves the highest performance across all metrics. As shown in Fig.~\ref{fig:ddb_ablation}-(d), applying all losses leads to clear extraction of moiré patterns in the pixel domain (Fig.~\ref{fig:ddb_ablation}-(d)-(i),(ii)) and, unlike Fig.~\ref{fig:ddb_ablation}-(c), achieves much better separation of moiré patterns and content in the feature domain (Fig.~\ref{fig:ddb_ablation}-(d)-(i),(ii)).

\subsubsection{Ablation Study on Moiré-adaptive Cross-attention Block (MCB)} \label{subsubsec_exp:mcb_ablation}

\begin{figure}[!t]
    \centerline{\includegraphics[width=\columnwidth]{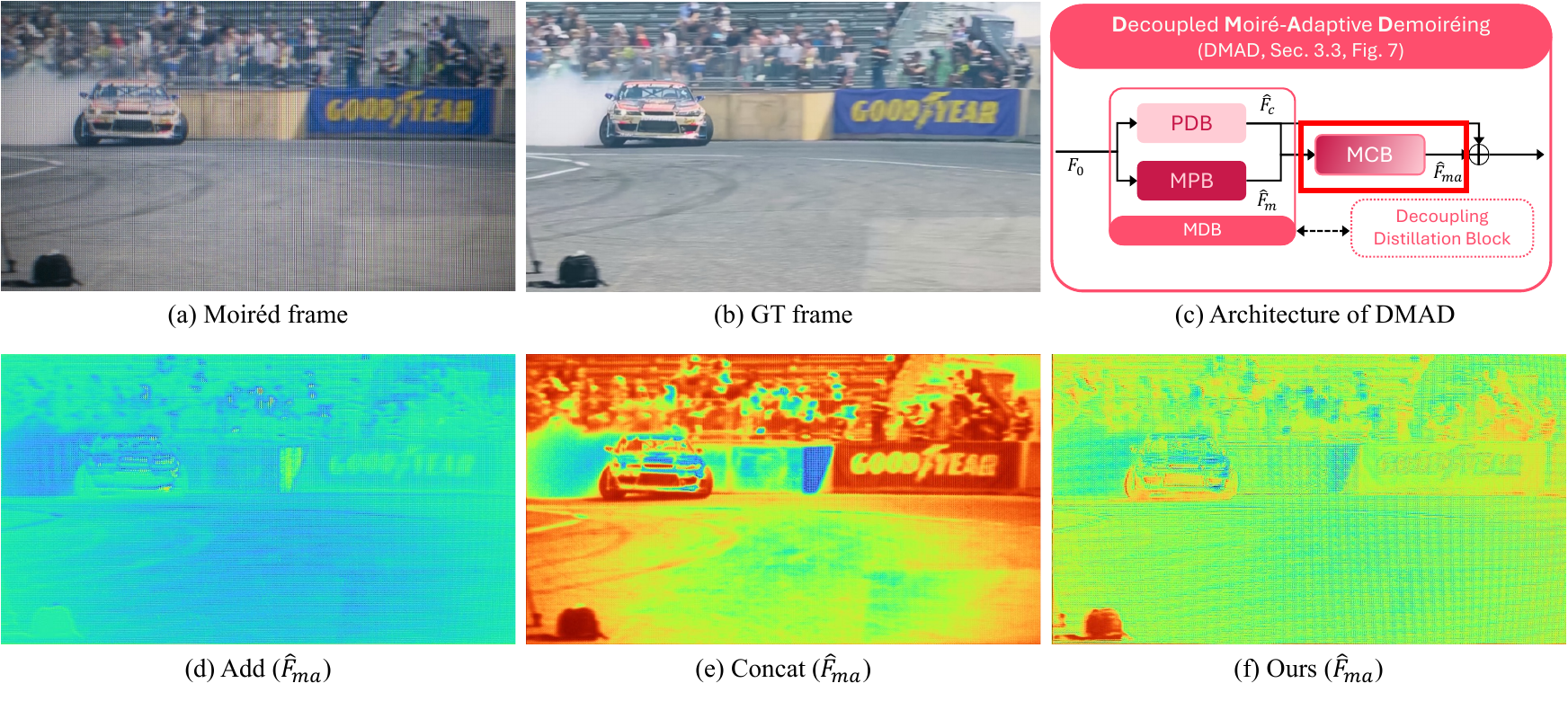}}
    \caption{Ablation study on the MCB module, comparing different feature fusion strategies for the separated moiré and content features. (a) and (b) show the input moiré frame and the corresponding ground-truth frame, respectively. (c) depicts the architecture of DMAD, where $\hat{F}_{ma}$ from (c) is used as the feature for visualization in (d)–(f). (d)–(f) visualize the fused feature maps using different fusion methods: addition (d), concatenation (e), and cross-attention (f). Compared to (d) and (e), the use of cross-attention in (f) results in more prominent moiré patterns, indicating that the fused features are more moiré-adaptive.}
    \label{fig:mcb_ablation}
\end{figure}

Table~\ref{tab:mcb_ablation} and Fig.~\ref{fig:mcb_ablation} present experiments comparing different operations for fusing the moiré and content features within the MCB. As shown in Table~\ref{tab:mcb_ablation}, performing cross-attention (Table~\ref{tab:mcb_ablation}-(c)) yields better performance than simply adding the two features (Table~\ref{tab:mcb_ablation}-(a)) or concatenating them (Table~\ref{tab:mcb_ablation}-(b)).
Fig.~\ref{fig:mcb_ablation} visualizes the resulting moiré-adaptive features generated by each operation. Consistent with the quantitative results, the cross-attention operation produces features that are more effectively adapted to moiré patterns compared to the add or concatenate operations (Fig.~\ref{fig:mcb_ablation}-(c),(d)).

\begin{table}
  \centering
  \begin{tabular}{@{}lccc@{}}           
    \toprule
    \textbf{Operation} & \textbf{PSNR}$\uparrow$ & \textbf{SSIM}$\uparrow$ & \textbf{LPIPS}$\downarrow$ \\ \midrule
    Addition              &                             28.3981  & \textcolor{blue}{\underline{0.9172}}& \textcolor{blue}{\underline{0.0969}} \\
    Concat                & \textcolor{blue}{\underline{28.4035}}&                             0.9170  &                             0.0970  \\
    Cross-Attention (Ours)&     \textcolor{red}{\textbf{28.5211}}&     \textcolor{red}{\textbf{0.9185}}&     \textcolor{red}{\textbf{0.0952}} \\
    \bottomrule
  \end{tabular}
  \caption{Performance according to fusion operation. Add means element-wise addition, Cat means concatenation and apply 1 $\times$ 1 conv, CA means Cross-Attention. We observe that applying the cross-attention operation yields the best performance compared to addition or concatenation.}
  \label{tab:mcb_ablation}
\end{table}

\subsubsection{Ablation Study on Alignment Module} \label{subsubsec_exp:alignment}

\begin{table}[!t]
  \centering
  \begin{tabular}{@{}lccccc@{}}           
    \toprule
    \textbf{Align} & \textbf{PSNR}$\uparrow$ & \textbf{SSIM}$\uparrow$ & \textbf{LPIPS}$\downarrow$ & \textbf{tOF}$\downarrow$ & \textbf{Params(M)} \\ \midrule
        (a) \textbf{Flow}~\cite{fmanet}-based&                            28.3764  &                            0.9180  &                            0.0975  &                            1.4078 & 2.35 \\
        (b) \textbf{DCN}~\cite{dcnv2}-based  &                            28.3791  &                            0.9162  &                            0.0984  &                            1.4212 & 2.91 \\
        (c) \textbf{PCD}~\cite{edvr}-based   &                            28.5055  &    \textcolor{red}{\textbf{0.9189}}&    \textcolor{red}{\textbf{0.0939}}&    \textcolor{red}{\textbf{1.3936}}& 5.28 \\
        (d) \textbf{Patch}~\cite{psrt}-based &\textcolor{blue}{\underline{28.5132}}&                            0.9184  &                            0.0958  &                            1.4223 & 3.78 \\
        (e) \textbf{Implicit (Ours)}     &    \textcolor{red}{\textbf{28.5211}}&\textcolor{blue}{\underline{0.9185}}&\textcolor{blue}{\underline{0.0952}}&\textcolor{blue}{\underline{1.4016}}& \textbf{2.34} \\
    \bottomrule
  \end{tabular}
  \caption{Ablation study on alignment methods. For quantitative evaluation of temporal consistency, we employ tOF \cite{tof}, a metric that measures temporal consistency. (a)–(d) apply explicit alignment modules to align three neighboring frames to the center frame, while (e) does not use any explicit alignment module. Specifically, (a) \textbf{Flow} employs flow-based alignment following the FMA-Net~\cite{fmanet} style, (b) \textbf{DCN} uses offset-based alignment with DCN~\cite{dcnv2}, (c) \textbf{PCD} adopts the PCD alignment from EDVR~\cite{edvr}, and (d) \textbf{Patch} applies the patch alignment proposed by Shi et al.~\cite{psrt}. As shown in the results, the implicit alignment approach used in \textbf{Ours} generally achieves better performance, with lower tOF and the highest parameter efficiency.}
  \label{tab:alignment}
\end{table}

\begin{figure}[!t]
    \centerline{\includegraphics[width=\columnwidth]{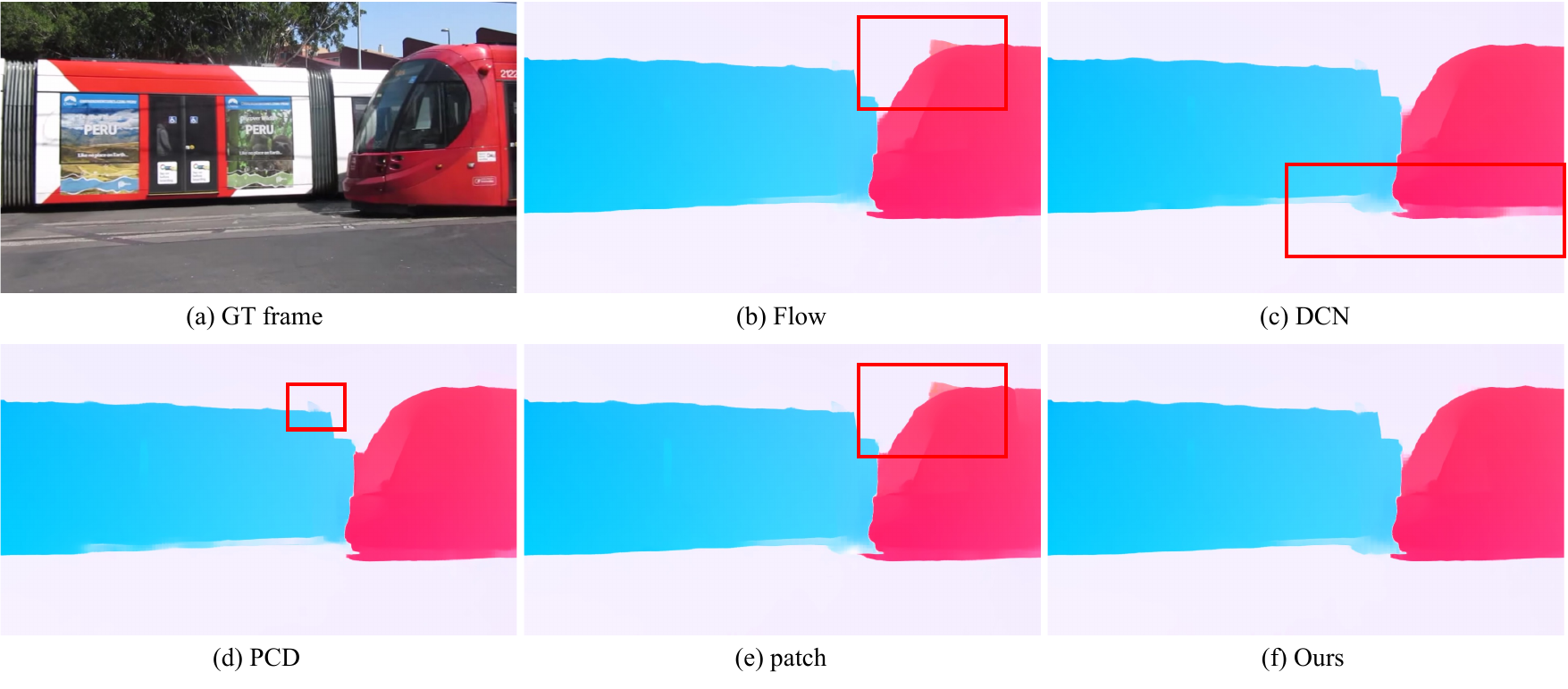}}
    \caption{Qualitative results of different alignment methods. To evaluate the temporal consistency of each model, we visualize the optical flow extracted after applying each alignment module. (b)–(e) represent explicit alignment methods, while (f) \textbf{Ours} adopts an implicit alignment strategy. Specifically, (b) \textbf{Flow} utilizes flow-based alignment following the FMA-Net~\cite{fmanet} style, (c) \textbf{DCN} applies offset-based alignment using DCN~\cite{dcnv2}, (d) \textbf{PCD} adopts PCD alignment from EDVR~\cite{edvr}, and (e) \textbf{Patch} uses the patch alignment proposed by Shi et al.~\cite{psrt}. As highlighted in the red boxes of (b)–(e), explicit alignment methods tend to produce noisy flow maps, indicating less stable temporal consistency.}
    \label{fig:align} 
\end{figure}

Table~\ref{tab:alignment} evaluates the alignment performance of STAD as described in Sec.~\ref{subsec_pm:stad}. The proposed STAD does not incorporate any explicit alignment module. This design choice stems from the fact that the complex characteristics of moiré patterns can make explicit alignment modules detrimental to the demoiréing task. To assess the effectiveness of our implicit alignment, we additionally evaluate versions of the model with explicit alignment modules and use tOF \cite{tof} as the metric to measure temporal consistency, alongside a qualitative assessment based on extracted optical flow.
In Table~\ref{tab:alignment}, (a) applies the optical flow-based alignment used in FMA-Net \cite{fmanet}, (b) employs DCN-based alignment \cite{dcnv2}, (c) integrates the PCD alignment proposed in EDVR \cite{edvr}, and (d) adopts the patch alignment method from Shi et al.~\cite{psrt}. Similar to previous alignment methods~\cite{psrt,vdemoire,rawvdemoire,dtnet,edvr}, the frame alignment module is inserted immediately after feature extraction, followed by the demoiréing process.
Results show that (a), (b), and (d) yield worse PSNR$\uparrow$, SSIM$\uparrow$, and LPIPS$\downarrow$ compared to (e). While (c) achieves comparable performance in all three metrics and tOF, it requires more than twice the number of parameters compared to (e). Furthermore, Fig. Fig.~\ref{fig:align}-(f) shows that the optical flow in Fig.~\ref{fig:align}-(b), Fig.~\ref{fig:align}-(c), and Fig.~\ref{fig:align}-(e) is noisier, particularly in the areas highlighted by the red boxes. These observations demonstrate that the implicit alignment in the proposed STAD is both efficient and effective for the video demoiréing task.

\subsubsection{Ablation Study on Training Strategy} \label{subsubsec_exp:training_strategy}

\begin{table}
  \centering

  \begin{tabular}{c cc ccc} 
    \toprule
    \multicolumn{1}{c}{} &
    \multicolumn{2}{c}{\textbf{Train Stage}} &
    \multicolumn{3}{c}{\textbf{Metrics}} \\
    \cmidrule(lr){2-3}\cmidrule(l){4-6}
        & stg1                                  & stg2 & PSNR$\uparrow$ & SSIM$\uparrow$ & LPIPS$\downarrow$ \\ \midrule
    (a) & \textcolor{red}{\textbf{D}}/\textcolor{red}{\textbf{S}} & -                   & 27.9926 & 0.9130 & 0.1002 \\
    (b) & \textcolor{red}{\textbf{S}} & \textcolor{red}{\textbf{D}}/\textcolor{blue}{\textbf{S}} & \textcolor{blue}{\underline{28.1729}}& \textcolor{blue}{\underline{0.9169}}& \textcolor{blue}{\underline{0.0995}} \\
    (c) & \textcolor{red}{\textbf{S}} & \textcolor{red}{\textbf{D}}/\textcolor{red}{\textbf{S}}  &     \textcolor{red}{\textbf{28.5211}}&     \textcolor{red}{\textbf{0.9185}}&     \textcolor{red}{\textbf{0.0952}} \\
    \bottomrule
  \end{tabular}
  
  \caption{Ablation study on training strategies. \textbf{D} and \textbf{S} denote DMAD and STAD, respectively. \textcolor{red}{Red text} indicates modules that are trained, while \textcolor{blue}{blue text} represents modules that are frozen. The results show that the strategy of first training STAD in stage 1, followed by jointly training DMAD and STAD in stage 2, achieves the best performance.}
  \label{tab:training_strategy}
\end{table}

As discussed in Sec.~\ref{subsec_pm:training}, we adopt a two-stage training strategy to effectively optimize the proposed MoCHA-former. To demonstrate the effectiveness of this approach, we compare different training strategies in Table~\ref{tab:training_strategy}. In this table, \textbf{D} and \textbf{S} denote the DMAD and STAD modules, respectively. Red text indicates that the corresponding module is being trained, while blue text indicates that the module is frozen. As shown in Table~\ref{tab:training_strategy}-(a), jointly optimizing both DMAD and STAD results in significant performance degradation compared to our method. The losses associated with the DDB for optimizing DMAD are designed to disentangle moiré patterns from content, while the loss functions used for STAD aim to remove moiré artifacts. These two objectives are inherently conflicting, leading to suboptimal performance when optimized jointly.

Table~\ref{tab:training_strategy}-(b) adopts a two-stage training strategy: in stage 1, only STAD is trained, and in stage 2, STAD is frozen while DMAD is trained. While this approach improves over joint optimization, its performance is still inferior to that of Table~\ref{tab:training_strategy}-(c). The rationale behind this design is to first perform general spatio-temporal video demoiréing using STAD, and subsequently train DMAD to adaptively remove residual moiré patterns. However, even if STAD is sufficiently trained in stage 1, fine-tuning is still required to align with the moiré-adaptive features provided by DMAD. Therefore, as shown in Table~\ref{tab:training_strategy}-(c), we propose that the optimal training strategy is to first train STAD alone in stage 1, followed by jointly optimizing both DMAD and STAD in stage 2.


%% file: sections/Conclusion.tex
\section{Conclusion}

In this work, we have presented \textbf{MoCHA-former}, a novel raw-domain video demoiréing framework that explicitly decouples moiré patterns from content and implicitly aligns frames via lightweight spatio–temporal transformers. We have centered our design on two novel components: the Decoupled Moiré-Adaptive Demoiréing (DMAD) module, which produces moiré-conditioned features, and the Spatio-Temporal Adaptive Demoiréing (STAD) module, which models long-range dependencies while respecting channel-wise statistics of moiré patterns. We have structured DMAD to comprise the MDB module, which separates moiré patterns from content; the DDB module, which facilitates this separation; and the MCB module, which fuses the separated content and moiré features. Through this process, DMAD has generated moiré-adaptive features. In addition, STAD has incorporated the proposed SFB, which includes FCA to leverage spatio–temporal cues and channel-wise information in the frequency domain, and has introduced WFB to capture the characteristics of moiré patterns in the Fourier domain. Moreover, we have provided an in-depth analysis of moiré patterns through qualitative and quantitative experiments, highlighting their unique characteristics. Finally, we have demonstrated state-of-the-art performance on the RawVDemoiré dataset, outperforming previous approaches.

%% file: sections/Acknowledgements.tex
\section{Acknowledgements}

Will be updated soon
